\documentclass{nle}

\usepackage{hyperref}
\usepackage{amsmath}
\usepackage{graphicx}
\usepackage{natbib}
\ifpdf%
\usepackage{epstopdf}%
\else%
\fi
\usepackage{caption}
\usepackage{subcaption}
\usepackage{float}
\usepackage[table,xcdraw]{xcolor}
\usepackage{multirow}
\usepackage{arydshln}
\newcommand{\norm}[1]{\left\lVert#1\right\rVert}

\begin{document}
\label{firstpage}

\lefttitle{}
\righttitle{Papagiannopoulou et al.}

\papertitle{Article}

\jnlPage{1}{00}
\jnlDoiYr{2019}
\doival{10.1017/xxxxx}

\bibliographystyle{nlelike.bst}

\title{Keywords lie far from the mean of all words in local vector space}

\begin{authgrp}
\author{Eirini Papagiannopoulou, Grigorios Tsoumakas and Apostolos N. Papadopoulos}
\affiliation{Aristotle University of Thessaloniki, 54124, Thessaloniki, Greece \\ 
        \email{\{epapagia, greg, papadopo\}@csd.auth.gr}}
\end{authgrp}

% \begin{authgrp}
% \author{Grigorios Tsoumakas}
% % \affiliation{Aristotle University of Thessaloniki, 54124, Thessaloniki, Greece
% %         \email{greg@csd.auth.gr}}
% \end{authgrp}

% \begin{authgrp}
% \author{Apostolos N. Papadopoulos}
% % \affiliation{Aristotle University of Thessaloniki, 54124, Thessaloniki, Greece
% %         \email{papadopo@csd.auth.gr}}
% \end{authgrp}

\history{(Received xx xxx xxx; revised xx xxx xxx; accepted xx xxx xxx)}
%\received{20 March 1995; revised 30 September 1998}

\begin{abstract}
Keyword extraction is an important document process that aims at finding a small set of terms that concisely describe a document's topics. The most popular state-of-the-art unsupervised approaches belong to the family of the graph-based methods that build a graph-of-words and use various centrality measures to score the nodes (candidate keywords). In this work, we follow a different path to detect the keywords from a text document by modeling the main distribution of the document's words using local word vector representations. Then, we rank the candidates based on their position in the text and the distance between the corresponding local vectors and the main distribution's center. We confirm the high performance of our approach compared to strong baselines and state-of-the-art unsupervised keyword extraction methods, through an extended experimental study, investigating the properties of the local representations.  
\end{abstract}

\maketitle

\section{Introduction}
\label{sec:intro}

Automatic Keyword Extraction (AKE) aims at finding the most representative words that summarize the contents of a document, without the expensive and time-consuming involvement of human annotators \citep{hasan+ng2014, DBLP:journals/ipm/Vega-OliverosGM19}. AKE is useful in several tasks, such as query expansion \citep{DBLP:conf/jcdl/SongSAO06}, faceted search \citep{gutwin1999improving}, text summarization \citep{DBLP:journals/wias/ZhangZM04}, document clustering and text classification \citep{DBLP:conf/acl/HulthM06}. As the amount of online digital textual information, e.g., e-newspapers, digital libraries with scientific literature, encyclopedias, blogs, magazines, etc., increases, the task of AKE becomes more and more popular, attracting much attention from the research community \citep{papagiannopoulou2019review, DBLP:journals/ipm/Vega-OliverosGM19, DBLP:journals/jiis/MerrouniFO20}. 

Several AKE methods have been proposed in the past.
Traditional unsupervised graph-based approaches consider the central nodes of a graph-of-words as the most representative ones \citep{mihalcea+tatau2004}. Moreover, there exist strong baselines that use common statistics (e.g. term frequency-inverse document frequency, known as Tf-Idf) and/or heuristics (e.g., position in the document) to detect the most significant terms. Furthermore, state-of-the-art approaches for the task span both classical supervised machine learning methods \citep{medelyan2009human} and deep learning techniques \citep{DBLP:conf/acl/MengZHHBC17} that perform better compared to unsupervised ones. However, supervised methods demand labelled data and are biased towards the domain of the training corpus. Another drawback of most AKE methods is the use of external tools, e.g., part-of-speech (POS) taggers, for grammatical/syntactic analysis and the use of (pre-trained) static or dynamic word embeddings that have a bias over the corpora domains used for training or need supervised fine-tuning.  

In this paper, inspired by the idea of using local word vector representations \citep{papagiannopoulou2018local} and the outlying behaviour of the keywords \citep{Papagiannopoulou2019outliers}, we propose a new unsupervised method for keyword extraction. First, our approach builds a local vector representation for each word that encodes the words' neighbourhood. Then, we estimate the \textit{distribution's center (mean vector) of the words}. Since the majority of a document's terms have neutral meaning or hardly related to the documents' topics, the main bulk of the words that determine the distribution's center are the non-keywords. Finally, we score and rank the words utilizing their corresponding distance from the distribution's center and the position of their first occurrence in the text (the index of the first sentence where the word occurs first), as keywords usually appear early in the document \citep{DBLP:conf/acl/FlorescuC17} and are distant from the distribution's center. Our empirical study confirms the high performance of our approach (in terms of the F$_1$-measure) on three datasets compared to state-of-the-art unsupervised keyword extraction methods.

\begin{figure}[H]
\centering
    \centering
    \includegraphics[width=0.75\linewidth]{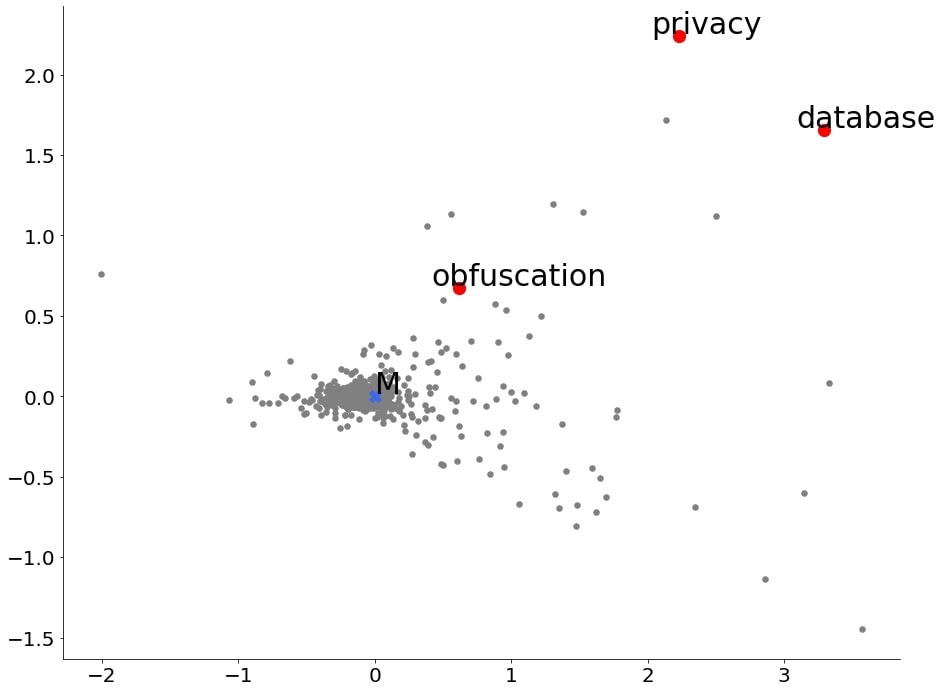}
  
  \caption{PCA 2d projection of the 50d local GloVe vectors for the article with title  ``Obfuscated databases and group privacy'' and keywords \{\textit{database}, \textit{obfuscation}, \textit{privacy}\}.}
  \label{fig:inspiration1}
\end{figure}

As a small motivational example, Fig.~\ref{fig:inspiration1} projects 50d local GloVe \citep{Pennington14glove:global} representations of the words learned by training GloVe on the full-text of a target academic publication from the NUS \citep{DBLP:conf/icadl/NguyenK07} collection on the first two principal components. The figure's caption provides information regarding the article's title and keywords. We notice that keywords are far from the main bulk of the words that are centered around (0,0). The center of the distribution is the blue ``x'' annotated with M (mean). Similar plots, supportive of our key intuition, are also obtained from other documents (see Fig.~\ref{fig:inspiration2} in the Appendix for an additional example).

Our contributions are the following: (1) We propose a new unsupervised AKE approach that employs local word vector representations to model the main distribution of a document's words. Counter to existing methods, e.g., Key2Vec \citep{DBLP:conf/naacl/MahataKSZ18}, EmbedRank \citep{DBLP:conf/conll/Bennani-SmiresM18}, RVA \citep{papagiannopoulou2018local}, etc., we consider the distribution's center closer to the non-keywords, since the main bulk of words are neutral or slightly relevant to the documents' topics. The proposed method outperforms strong baselines and state-of-the-art approaches of the task. (2) We show via an empirical study that simpler local word vector representations, based on the words' co-occurrences within a specific window, encode statistical information equivalent to the one encoded by the local GloVe word embeddings. (3) We investigate the properties of local word embeddings and the effect of robust statistics in the calculation of the distribution's center. Unlike the main direction in text mining that uses the cosine similarity/distance, we suggest distance metrics that incorporate the vectors' magnitude or covariance estimations to capture the similar behaviour of the non-keywords' vectors over the keywords' ones.

The rest of the paper is organized as follows. Section \ref{sec:related_work} presents related work on  unsupervised AKE (Section \ref{subsec:AKE}) and  pre-trained/local embeddings in the context of this task (Section \ref{subsec:embeddings}). Section \ref{sec:our_approach} describes our approach (Section \ref{subsec:description}) providing intuitions regarding its relationship with the graph-based approaches (Section \ref{subsec:relationship_with_graph_based_methods}). Section \ref{sec:experiments} investigates the properties of the local representations (Section \ref{subsec:tuning}), gives a comparison between our approach and the other state-of-the-art unsupervised keyword extraction methods (Section \ref{subsec:comparison}) and provides a qualitative study (Section \ref{subsec:qualitative_results}). Finally, Section \ref{sec:conclusions} concludes our work and presents future research directions.

\section{Related Work}
\label{sec:related_work}

Keyword extraction is important for many tasks in text processing/mining (see Section \ref{sec:intro}). Moreover, it is the core process for various keyphrase extraction methods that form and rank the candidate phrases using the already scored candidate unigrams by a keyword extraction approach \citep{wan+xiao2008, DBLP:conf/acl/FlorescuC17}. Furthermore, word embeddings came to the foreground by \cite{DBLP:journals/corr/abs-1301-3781}, with the Continuous Bag-of-Words model (CBOW) and the Continuous Skip-gram model. Since then, many AKE methods utilise word embeddings, pre-trained or local ones, as either an external semantic information source or a more expressive representation than plain statistics, respectively. This section presents related work on the unsupervised AKE as well as the different types of word embeddings used in the context of this task.

\subsection{Unsupervised Keyword Extraction}
\label{subsec:AKE}

TextRank \citep{mihalcea+tatau2004} inspired researchers to build popular state-of-the-art methods. TextRank keeps only nouns and adjectives as candidates. Then, it constructs an undirected and unweighted graph-of-words with the candidates as nodes and edges between the nodes that co-occur within a window of W words. Finally, the PageRank \citep{grin+page1998} algorithm runs until it converges. SingleRank \citep{wan+xiao2008} is an extension of TextRank which incorporates weights to edges, i.e., each edge weight is equal to the number of co-occurrences of the two corresponding words within a specific window. In this vein, RAKE \citep{rose2010automatic} creates a graph of word-word co-occurrences and assigns as a score to each candidate word its frequency or degree or the ratio of degree to frequency. Later, \cite{DBLP:conf/acl/FlorescuC17} proposed PositionRank, a graph-based unsupervised method that not only considers the word-word co-occurrences but also incorporates each word's positions through a biased PageRank. \cite{DBLP:conf/ecir/RousseauV15} propose $k$-Core decomposition \citep{seidman1983network, DBLP:journals/corr/cs-DS-0202039, DBLP:journals/corr/cs-DS-0310049} of unweighted or weighted graphs-of-words, whereas \cite{DBLP:conf/emnlp/TixierMV16} apply the k-truss \citep{cohen2008trusses} algorithm to the task of keyword extraction, a triangle-based extension of $k$-Core. Recently, \cite{DBLP:journals/eswa/BiswasBS18} proposed a graph-based keyword extraction method from tweets, using collective node weight that depends on many parameters, e.g., node's centrality and strength of neighbours \citep{barrat2004architecture} as well as word's frequency and position.

\cite{DBLP:journals/ipm/Vega-OliverosGM19} present the performance of many centrality measures in terms of precision, recall and F$_1$-score, suggesting the use of a combination of word rankings that the various centrality measures return, based on a multi-centrality index. Their experimental study includes nine different centrality measures and three collections. The centrality measures in graph-of-words considered for AKE are Eccentricity, Eigenvector, Clustering Coefficient, Betweenness, Degree, Closeness, PageRank, Structural Holes, $k$-Core. Although all measures show comparable results, PageRank performs better in most cases. Earlier, \cite{DBLP:journals/corr/LahiriCC14} show that simple centrality measures, e.g.,  Degree and Strength (weighted Degree), or statistics, such as term frequency (Tf) and term frequency-inverse document frequency (Tf-Idf), perform as well as PageRank. Finally, recent statistical approaches, such as YAKE \citep{DBLP:conf/ecir/0001MPJNJ18, DBLP:journals/isci/CamposMPJNJ20}, use statistical metrics that capture context information and the spread of the terms into the document, besides the term's position/frequency, etc.

\subsection{Pre-trained and Local Word Embeddings}
\label{subsec:embeddings}

Various types of word embeddings are used in the context of \textit{keyphrase} extraction. First, \cite{Wang2014, DBLP:conf/adc/WangLM15} proposed graph-based approaches that incorporate semantic information coming from distributed word representations. Particularly, they have experimented with several pre-trained publicly available word vectors: SENNA \citep{DBLP:journals/jmlr/CollobertWBKKK11}, Turian's \citep{DBLP:conf/acl/TurianRB10} and HLBL \citep{DBLP:conf/icml/MnihH07} embeddings (the last one achieves the highest performance). This approach motivated \cite{DBLP:conf/naacl/MahataKSZ18} to present Key2Vec, a graph-based keyphrase extraction method that uses Fasttext \citep{DBLP:journals/tacl/BojanowskiGJM17, DBLP:conf/eacl/GraveMJB17} (multi-)word embeddings (trained on a corpus of scientific abstracts) to represent the candidate keyphrases of a document.

Later, EmbedRank \citep{DBLP:conf/conll/Bennani-SmiresM18} employed publicly available pre-trained sentence embeddings, Doc2Vec \citep{DBLP:conf/rep4nlp/LauB16} or Sent2Vec \citep{DBLP:conf/naacl/PagliardiniGJ18}, to embed both candidate phrases and documents in the same high-dimensional vector space. Finally, the system ranks the candidate phrases using the cosine similarity between the embedding of the candidate phrase and the document embedding. In this vein, \cite{papagiannopoulou2018local} present the Reference Vector Algorithm (RVA), a method for keyphrase extraction, whose main innovation is the use of local GloVe \citep{Pennington14glove:global} word vectors trained only on the target document  (experiments with pre-trained embeddings lead to low performance). 

\section{Our Approach}
\label{sec:our_approach}

We first describe our approach in Section \ref{subsec:description}. Then, we provide intuitions regarding the relationship between the proposed method and the graph-based approaches in Section \ref{subsec:relationship_with_graph_based_methods}.

\subsection{Description}
\label{subsec:description}

In the following four subsections, we present the four steps (stages) of our method: (a) the pre-processing of the document, (b) the formation of the local word vectors, (c) the calculation of the main distribution's center, (d) the scoring and ranking process of the candidate keywords.

\subsubsection{Pre-processing of the Document}
\label{subsub:preprocessing}

Following the paradigm of graph-based approaches \citep{mihalcea+tatau2004, wan+xiao2008, DBLP:conf/acl/FlorescuC17}, where syntactic filters often restrict the candidate terms to nouns and adjectives, we remove from the target document: punctuation marks, stopwords, words with less than two characters and tokens consisting only of digits. Apart from the classic stopwords list of the English language, we also remove common adjectives, reporting verbs, determiners and functional words\footnote{All the corresponding lists of words are available on our GitHub repository (URLs are provided in Section \ref{subsec:setup}).}. This way, we avoid considering trivial or unimportant terms that are unlikely to be keywords. Then, we apply stemming to reduce the inflected word forms into the root ones, ending up to the final set of $n$ candidates $W (w \in W)$. 

\subsubsection{Local Vector Representations}
\label{subsub:word_vectors}

\cite{papagiannopoulou2018local} show that local word embeddings perform better in the keyphrase extraction task than the global ones, i.e., pre-trained word vectors or vectors trained on large collection(s). Such local word vector representations came from the training of the GloVe model on the target document. This way, the ostensibly independent document's statistics are encoded in a more expressive vector representation, as the GloVe model creates word vectors such that the dot product of each pair of vectors $(w_i, \tilde{w}_k)$ equals the logarithm of the probability of co-occurrence of the corresponding words in the text ($X_{ik}$), i.e., 
\begin{equation}
w_i^T\tilde{w}_k + b_i + \tilde{b}_k = log(X_{ik})
\label{eq:glove_model}
\end{equation}
where $i, k$ comprise the given pair of words, $w \in ${\rm I\!R}$^d$ are the word vectors and $\tilde{w} \in ${\rm I\!R}$^d$ are the separate context word vectors, respectively, whereas, $b_i$ and $b_j$ are the corresponding bias terms (for further details regarding the GloVe technique and the weighted least squares regression model that proposes see the work of \cite{Pennington14glove:global}). 

Such type of statistics, e.g. the word-word co-occurrences, are also the primary source of information for graph-based keyword extraction methods. In this vein, the employed local training of GloVe on a single document and the graph-based family of methods can be considered as two alternative views of the same information source. In this work, we investigate the effectiveness of two different types of local word embeddings in the context of the AKE that use the document's source statistics to construct the corresponding vectors. Specifically, we have studied the following vector representations:
\begin{itemize}
    \item[i] local GloVe vector representations, i.e., the GloVe vectors' matrix $G_{n \times q}$  where $n$ is the number of unique stemmed words in the text and $q$ is the number of vectors' dimensions.
    
    \item[ii] word-word co-occurrence statistics, i.e., the term-term matrix $C_{n \times n}$  where $n$ is the number of unique stemmed words in the text and the $C_{x,j}$ element of the matrix contains the number of co-occurrences between the words $(w_x, w_j)$ ($1 \leq x \leq n$ and $1 \leq j \leq n$) within a window of 10 words (we express each word in terms of their co-occurrences with the rest of the words).
\end{itemize}

\subsubsection{Estimating the Center of the Main Distribution}
\label{subsub:distribution_location}

The next step of our approach is the estimation of the distribution's center (mean vector) of the document's local word vectors (each word participates once in the computation). We investigate the efficacy of two distinct estimations. The first one considers the sample's ($S$) estimated mean $\vec{\mu}_S$ of the corresponding vector matrix, i.e., the $G_{n \times q}$ or  $C_{n \times n}$, as the distribution's center. 

The second one employs the fast algorithm of \cite{DBLP:journals/technometrics/RousseeuwD99} for the MCD estimator \citep{Rousseeuw1984LeastRegression} to model the dominant distribution of the words' vectors. 
The MCD approach aims to find the subset $J$ of the word vector representations with size $h$ and mean $\vec{\mu}_J$ 
whose covariance matrix $\Sigma_J$ has the smallest determinant; 
\begin{equation}
    (\vec{\mu}_J, \Sigma_J): \mathrm{det}~\Sigma_J \leq \mathrm{det}~\Sigma_K\textrm{, }\forall K\subset D, |K|=h
\label{eq:mcd_estimator}
\end{equation}
where $\vec{\mu}_J$ and $\Sigma_J$ the robust distribution's center and covariance matrix of the corresponding word vectors' matrix, i.e., the $G_{n \times q}$ or  $C_{n \times n}$, respectively. This method, however, may break or not perform well in high-dimensional settings. For this reason, we apply the statistical procedure of Principal Component Analysis (PCA) \citep{pearson1901liii, hotelling1933analysis}, which uses an orthogonal transformation to convert a set of observations of possibly correlated variables into a set of values of linearly uncorrelated variables called principal components. This way, we keep only the first $m$ principal components, where $m$ is a small number, that finds the $m$ dimensions through the high-dimensional dataset in which the data is most spread out. 

Notice that our approach does not have to consider further term frequency thresholds or syntactic information, e.g., POS filters/patterns, for the candidate keyword identification. This means that there are no dependencies on external tools (e.g. POS taggers), which are also biased towards the domains of the corpora used for training. 

\subsubsection{Scoring Candidate Keywords}
\label{subsub:scoring}

We score each word with a value $S(w)$, $w \in W$ considering positional and context information according to the following formula: 
\begin{equation}
    S(w) = d(\vec{\mu}, \vec{v}) \times \frac{1}{z}
\label{eq:scoring_function}
\end{equation}
 where $d(\vec{\mu}, \vec{v})$ is the distance between the distribution's center $\vec{\mu}$ (i.e., $\vec{\mu}_S$ or $\vec{\mu}_J$) and the vector $\vec{v}$ of the word $w$ using the vector representation of our choice, i.e., GloVe vectors ($G_{n \times q}$) or term-term co-occurrences ($C_{n \times n}$),
respectively. The $z$ is the index of the first sentence where $w$ occurs in the document, i.e., $z = 1, 2, \dots, r$ with $r$ the total number of sentences. We choose not to use the word's index but only the sentence index, as it offers a more general/useful information regarding the word's importance. The higher the score $S(w)$, the more important the word $w$ for the document, i.e., we are interested in words that appear early in the document and have high distance values, as the main bulk of the words that determine the distribution's center are the non-keywords.

For the computation of the distance $d(\vec{\mu}, \vec{v})$, we use the Euclidean distance $d_E$, i.e.,
\begin{equation}
    d_E(\vec{v}, \vec{\mu}) = \sqrt{\sum_{i=1}^{\l}(v_i - \mu_i)^2}
\label{eq:euclidean_metric}
\end{equation}
where $l$ is the number of dimensions, i.e., equal to $q$, $n$ or $m$, depending on the type of the local vector representation and the need to apply dimensionality reduction with PCA or not. We have also experimented with the cosine and Mahalonobis distance measures (see Section \ref{subsec:tuning} for an extensive discussion on the various types of distance metrics and local vectors as well as the role of robust statistics in the corresponding calculations).

\subsection{Intuitions for the Relationship with the Graph-based Approaches}
\label{subsec:relationship_with_graph_based_methods}

The majority of unsupervised AKE graph-based approaches utilise statistics of the target document, e.g., word-word co-occurrences, words' position, etc. Such methods, first, construct a graph-of-words and, then, identify the most central vertices, considering them as keywords, by using various centrality measures, e.g., PageRank, Closeness, Betweenness, Eigenvector, $k$-Core, etc. In this section, we use the $k$-Core\footnote{We use the $k$-Core of undirected weighted graph-of-words constructed with a fixed-size sliding window equal to 10 following the pre-processing steps described in Section \ref{subsub:preprocessing}.} to provide intuitions concerning the relationship between the graph-based approaches and the proposed one, as it enables us to concentrate on the main goal of the study. Particularly, $k$-Core returns only the most cohesive connected component-s of the graph-of-words without scoring the corresponding graph nodes. Hence, there is no need to set a threshold for the number of the top N high-scored words that the measure will return (e.g. the case of PageRank). Other centrality measures offer similar visualizations.

The $k$-Core of a graph $G=(V, E)$ is the maximal subgraph that contains vertices of degree $k$ or more, where $V$ is the set of vertices and $E$ the set of edges. Thus, the $k$-Core of a graph-of-words comprises usually a \textit{minority} of ``qualified'' representatives for the whole graph (i.e., candidate keywords). Table \ref{tbl:k-core_statistics} presents the average number of vertices that belong to the k-Core ($|$k-Core$|$), the average total number of vertices $|V|$ and the average ratio $\frac{|k-Core|}{|V|}$ for the corresponding documents in the training sets of 3 different datasets, Semeval \citep{Kim:semeval2010}, Krapivin \citep{krapivin2009}, and NUS. We see that the k-Core subgraph in most cases contains a small percentage of nodes compared to $V$. This fact inspired us to think the document's important terms as a minority that differs from the main bulk of insignificant words, which are close to the distribution's center of the document's local word vectors.

\begin{table}[H]
\centering
\small
    \caption{$|k$-$Core|$ is the average number of vertices in the $k$-$Core$, $|V|$ the average total number of vertices and $\frac{|k-Core|}{|V|}$ is the average ratio for the documents of three datasets (Semeval, Krapivin, NUS).}
    \begin{tabular}{|c|c|c|c|}
    \hline
     Dataset & $|k$-$Core|$ & $|V|$ & $\frac{|k-Core|}{|V|}$ \\ \hline
    Semeval  &   70    &  645   &    0.110   \\ %\hline
    Krapivin &   74    &  736   &    0.103   \\ %\hline
    NUS      &   72    &  778   &    0.095   \\ \hline
    \end{tabular}

    \label{tbl:k-core_statistics}
\end{table}

\begin{figure}[H]
\centering
\begin{subfigure}{\linewidth}
  \centering
  \includegraphics[width=0.60\linewidth]{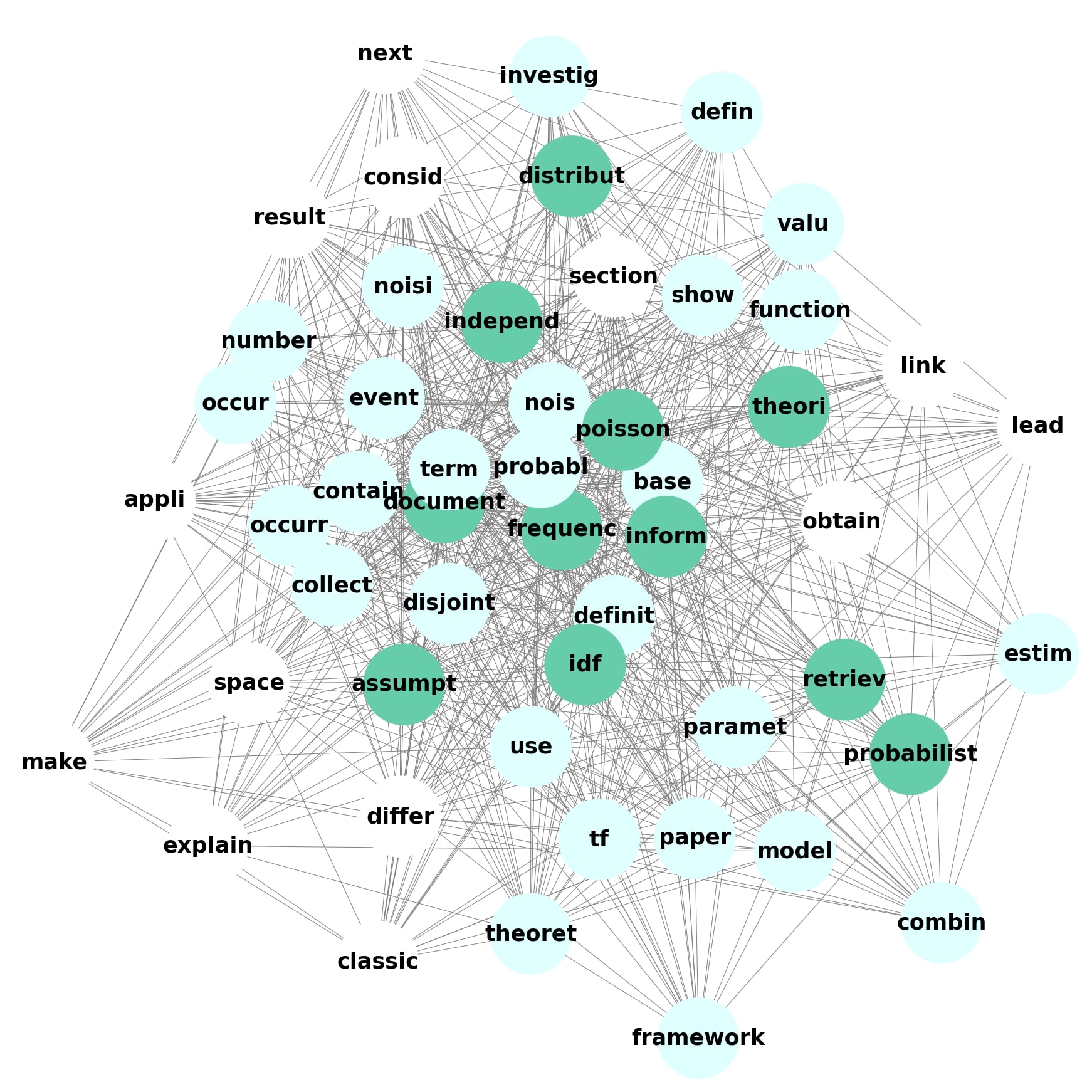}
  \caption{\textit{Title}: ``A frequency-based and a Poisson-based definition of the
probability of being informative''. \textit{Keywords}: inverse document frequency (idf), independence assumption, probabilistic information retrieval, poisson distribution, information theory
 }
  \label{fig:k-core_LVR_intersection_NUS_10}
\end{subfigure}

\begin{subfigure}{\linewidth}
  \centering
  \includegraphics[width=0.60\linewidth]{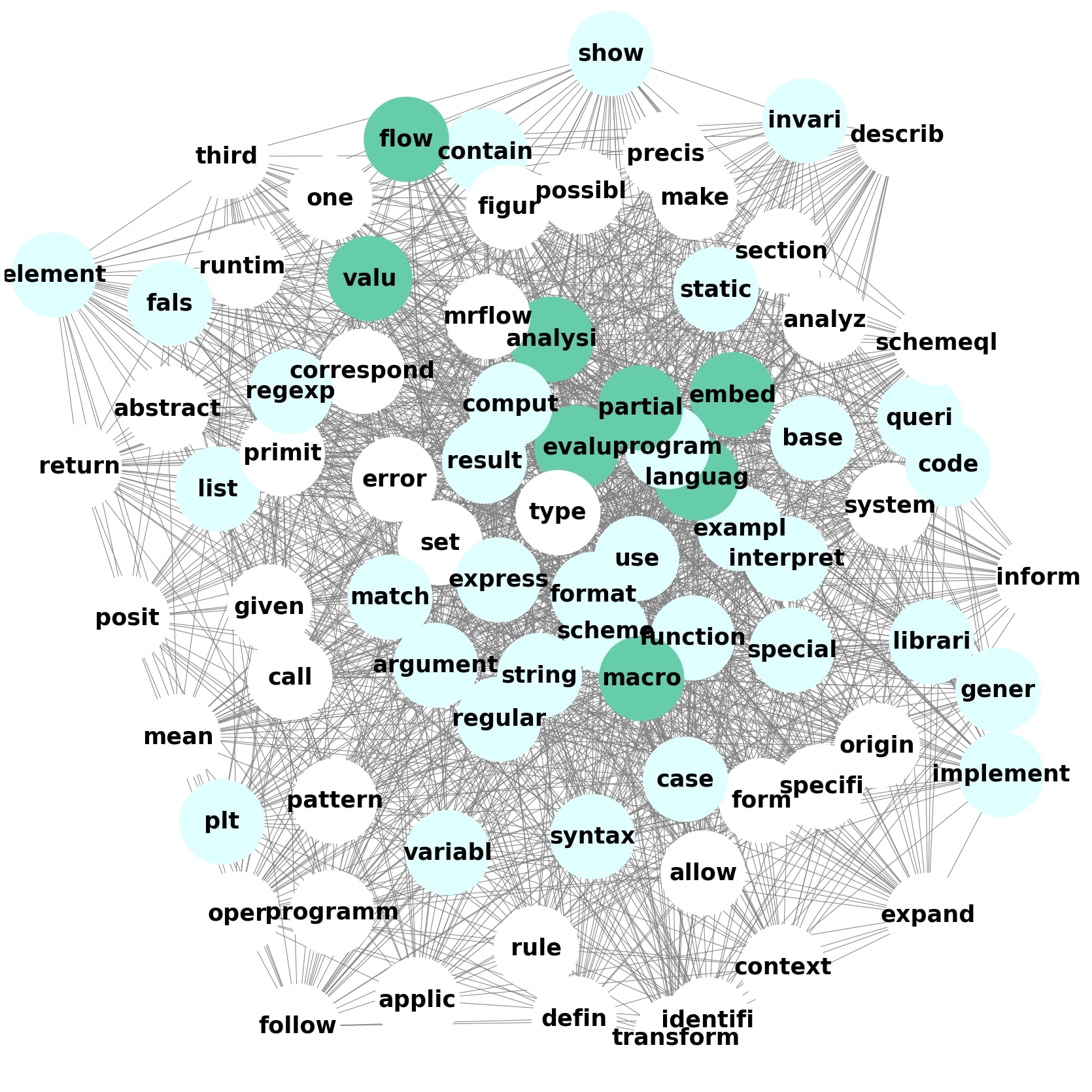}
  \caption{\textit{Title}: ``Improving the static analysis of embedded languages via partial evaluation''. \textit{Keywords}: partial, evaluation, macros, value flow analysis, embedded languages}
  \label{fig:k-core_LVR_intersection_NUS_107}
\end{subfigure}
\caption{$k$-Core sets for two articles.}
\label{fig:k-core_LVR_intersection_NUS}
\end{figure}

Furthermore, to get a feel of the similarities between the proposed method and the graph-based approaches, Figures \ref{fig:k-core_LVR_intersection_NUS_10} and \ref{fig:k-core_LVR_intersection_NUS_107} visualize the $k$-Core subgraph of two documents from the NUS training set. The coloured vertices represent the intersection of the top N = $|$k-Core$|$ candidate keywords returned by our method and the $k$-Core subgraph of the corresponding document, whereas the darker ones belong to both the intersection and the set of authors' keywords. Both approaches use the stemmed unigrams. Moreover, the caption of each figure provides information regarding the article's title and authors' keywords.

\section{Experimental Study}
\label{sec:experiments}

Section \ref{subsec:setup} presents the experimental setup of the study. In Section \ref{subsec:tuning}, we investigate the properties (behaviour) of the local vector representations in the context of the AKE task. Section \ref{subsec:comparison} shows a comparison of the proposed approach with other baselines and state-of-the-art methods, whereas Section \ref{subsec:qualitative_results} provides qualitative results.  

\subsection{Experimental Setup}
\label{subsec:setup}

We choose three well-known datasets that contain full-text articles from the computer science domain, NUS \citep{DBLP:conf/icadl/NguyenK07}, Semeval \citep{Kim:semeval2010} and Krapivin \citep{krapivin2009} datasets with 211, 244 and 2304 documents, respectively. Semeval is already divided into training (144) and test (100) sets. However, there are no explicit guidelines for the division of the Krapivin and NUS. Thus, we select the first 400 papers from the Krapivin (following the paradigm of \cite{DBLP:conf/acl/MengZHHBC17}) and the last 100 papers from NUS in alphabetical order as the testing data. We should clarify that the empirical study presented in Section \ref{subsec:tuning} uses only the training sets of the corresponding datasets, whereas the experimental comparison of Section \ref{subsec:comparison} employs the test sets. 

We compute F$_1$@5, F$_1$@10, F$_1$@15, as accuracy at the top of the ranking is more important in typical applications. We use exact string matching to determine the number of correctly matched words with the golden ones for a document. We also apply stemming to the output of the methods and the article's golden words, as a pre-processing step before the evaluation process. We employ the authors' keywords as a gold evaluation standard for all dataset collections. Regarding the GloVe setup, we used the default parameters (x$_{max}$ = 100, $\alpha = \frac{3}{4}$, window size = 10) \citep{Pennington14glove:global}, whereas the window for counting the co-occurrences between two words for the term-term word vector representation is equal to 10. The document frequency used by Tf-Idf approach (Section \ref{subsec:comparison}) is calculated separately for each dataset collection. The code of our method is available on our GitHub repository\footnote{\url{https://github.com/epapagia/LocalVectors_AKE}}, whereas, for the implementation of the competitive approaches, we employ the PKE\footnote{\url{https://github.com/boudinfl/pke}} toolkit and the NetworkX\footnote{\url{https://networkx.github.io/}} python library.

\subsection{Empirical Study on Local Vector Representations}
\label{subsec:tuning}

In this section, we use the training sets of the two smaller datasets (NUS and Semeval) to investigate the effectiveness of (a) the two local vector representations, i.e., GloVe and term-term co-occurrences vectors (t-t), (b) different distance metrics employed by the proposed approach via its scoring function (Equation \ref{eq:scoring_function}), (c) the robust statistics in the calculation of the distribution's center. Table \ref{tbl:results_trainset_custom_dist} presents the F$_1$-scores of the proposed approach on the two datasets using the two different types of word vectors. For the distance calculation, which is necessary to score the candidates, we use either the sample's mean (SM) or the robust mean (RM) given by the MCD estimator. We have experimented with various distance metrics including the Euclidean Distance (ED) (see Equation \ref{eq:euclidean_metric}), i.e.:
\begin{itemize}
    \item [i] Cosine Distance (CD):
    \begin{equation}
         d_C(\vec{v}, \vec{\mu}) = 1 - \frac{\vec{v} \cdot \vec{\mu}}{\norm{\vec{v}}_2 \norm{\vec{\mu}}_2}
    \label{eq:cosine_distance}
    \end{equation}
    
    \item [ii] Mahalanobis Distance (MD):
    \begin{equation}
         d_M(\vec{v},\vec{\mu}) = \sqrt{(\vec{v} - \vec{\mu})\Sigma^{-1}(\vec{v} - \vec{\mu})^T}
    \label{eq:mahalanobis_distance}
    \end{equation}
\end{itemize}
 where $d_C$ (or $d_M$) is the cosine (or Mahalanobis) distance between the distribution's center $\vec{\mu}$ (i.e., the sample's mean $\vec{\mu}_S$ or the robust mean $\vec{\mu}_J$ from the MCD estimator) and the vector $\vec{v}$ of the word $w$ using the vector representation of our choice, i.e., GloVe vectors ($G_{n \times q}$) or term-term co-occurrences ($C_{n \times n}$), respectively. In the case of the MD, $\Sigma$ represents the covariance matrix that can be (a) the sample's covariance (SC) matrix $\Sigma_S$, (b) the covariance matrix $\Sigma_{ML}$ based on the maximum likelihood covariance estimator \citep{rossi2018mathematical} (MLC) or (c) the robust covariance (RC) matrix $\Sigma_{J}$ from the MCD estimator. However, the MCD estimator may break or not perform well in high-dimensional settings. For this reason, we apply PCA to the GloVe and t-t vectors' matrices to keep only the first $m$ principal components, where $m$ is equal to 10.
 
 \begin{table}[H]
\centering
\small
\caption{F$_1$-scores of the proposed approach on the training sets of NUS and Semeval datasets using term-term co-occurrences vectors (t-t) and GloVe vectors (GloVe).}
\begin{tabular}{|l|l|l|l|l|}
\hline
\multicolumn{1}{|c|}{{\color[HTML]{000000} F$_1$@10}} & \multicolumn{2}{c|}{{\color[HTML]{000000} NUS}}       & \multicolumn{2}{c|}{{\color[HTML]{000000} Semeval}}   \\ \hline
Metric/Estimation type                             & \multicolumn{1}{c|}{t-t} & \multicolumn{1}{c|}{GloVe} & \multicolumn{1}{c|}{t-t} & \multicolumn{1}{c|}{GloVe} \\ \hline
ED/SM                                              & \textbf{0.452}           & 0.442                      & 0.388                    & 0.372                      \\
ED/RM (10-dim vecs)                                & 0.444                    & 0.443                      & 0.383                    & 0.376                      \\
CD/SM                                              & 0.347                    & 0.408                      & 0.320                    & 0.351                      \\
CD/RM (10-dim vecs)                                & 0.413                    & 0.407                      & 0.372                    & 0.353                      \\
MD/SM-SC                                           & 0.441                    & 0.435                      & \textbf{0.392}           & 0.382                      \\
MD/SM-MLC                                          & 0.441                    & 0.437                      & 0.391                    & 0.384                      \\
MD/RM-RC (10-dim vecs)                             & 0.424                    & 0.442                      & 0.363                    & 0.386                      \\ \hline
\end{tabular}

\label{tbl:results_trainset_custom_dist}
\end{table}

Table \ref{tbl:results_trainset_custom_dist} shows that t-t vectors in both datasets achieve the highest F$_1$-scores combined with Euclidean and Mahalanobis distance metrics, i.e., measures that consider the magnitude of the vectors and the correlations of the data, respectively. Only in the case of the cosine distance metric that is a judgment of orientation and not magnitude, GloVe vectors outperform the t-t vectors. In general, for all types of vectors, the scoring functions that incorporate Euclidean and/or Mahalanobis distance metrics achieve more or less equivalent results with high values compared to those that use the cosine distance.
 
Moreover, for the t-t vectors, the robust estimation of the sample's mean does not lead to improved performance (at least on the averages of the F$_1$-scores) except for the case of the cosine distance (once again the improvements are not enough to exceed the high F$_1$-score values achieved by the Euclidean and the Mahalanobis metrics). In this vein, the maximum likelihood and the robust covariance estimation that the Mahalanobis distance incorporates causes no improvement on the scores. Regarding the GloVe vectors, the robust estimation of the distribution's center, in most cases, slightly improves the method's performance. Once more, the maximum likelihood and the robust covariance estimations within the Mahalanobis distance increase the corresponding F$_1$-scores.

\begin{figure}[H]
\centering

  \begin{subfigure}{0.75\linewidth}
    \centering\includegraphics[width=1.0\linewidth]{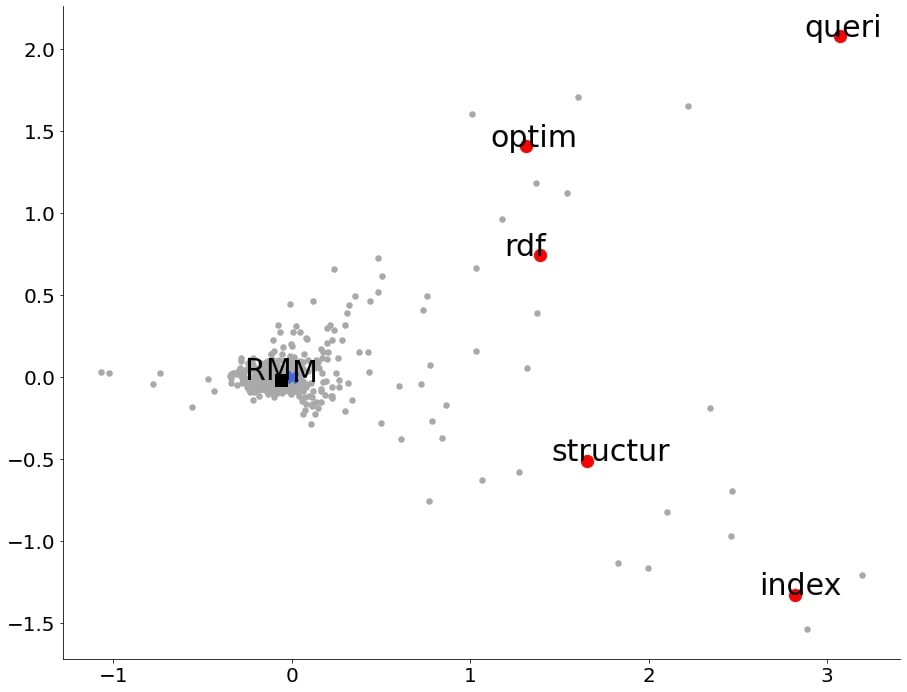}
    \caption{}
  \end{subfigure}
  
  \begin{subfigure}{0.75\linewidth}
    \centering\includegraphics[width=1.0\linewidth]{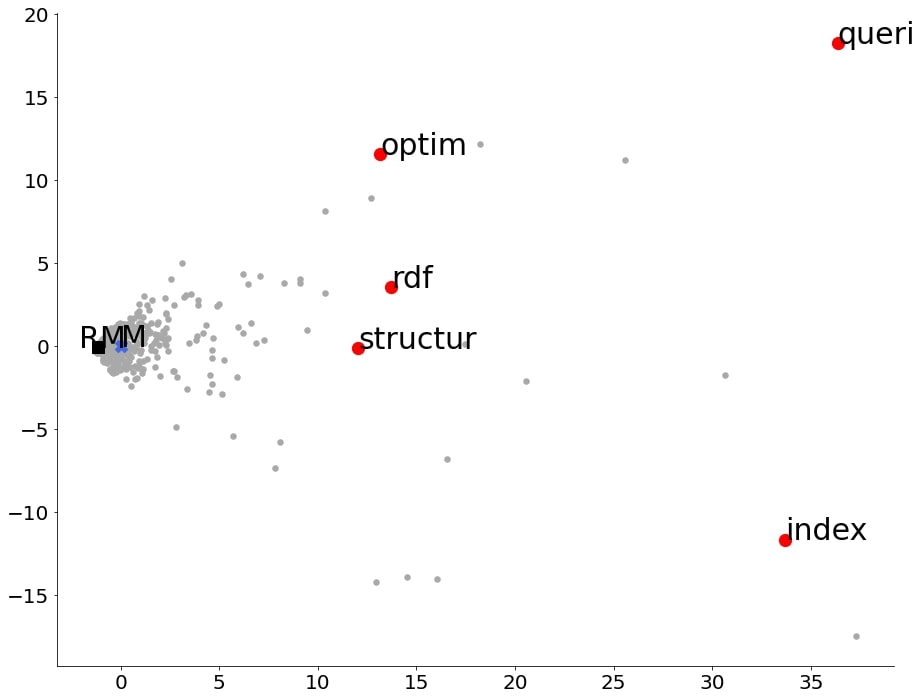}
    \caption{}
  \end{subfigure}
  
  \caption{PCA 2d projection of the (a) 50d local GloVe and (b) term-term local embeddings for the article.}
  \label{fig:NUS_109_unnormalized_local_embeddings}
\end{figure}

\begin{figure}[H]
  \centering
  \begin{subfigure}{0.75\linewidth}
    \centering\includegraphics[width=1.0\linewidth]{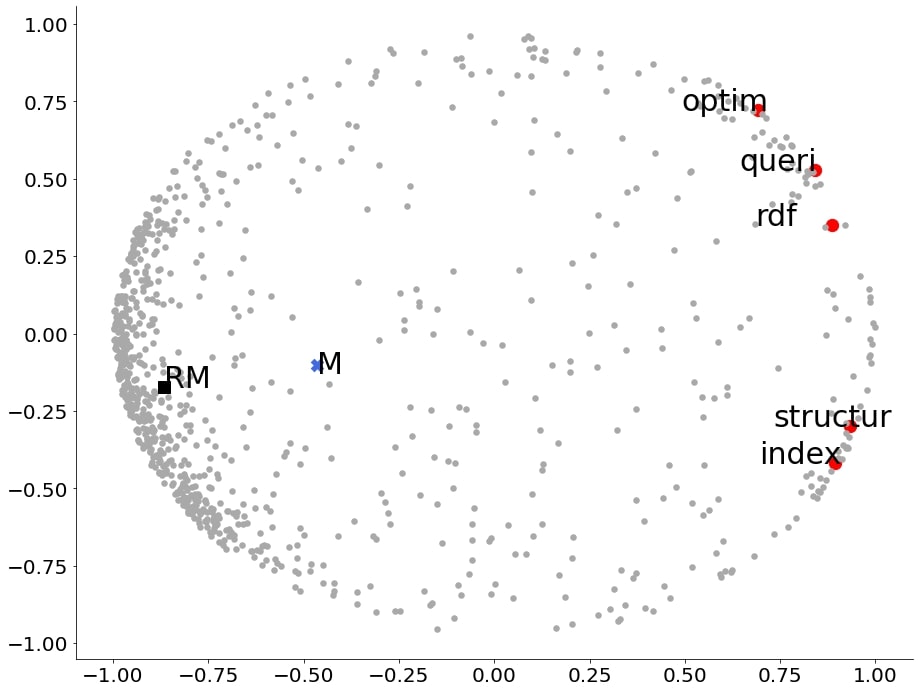}
    \caption{}
  \end{subfigure}
 
  \begin{subfigure}{0.75\linewidth}
    \centering\includegraphics[width=1.0\linewidth]{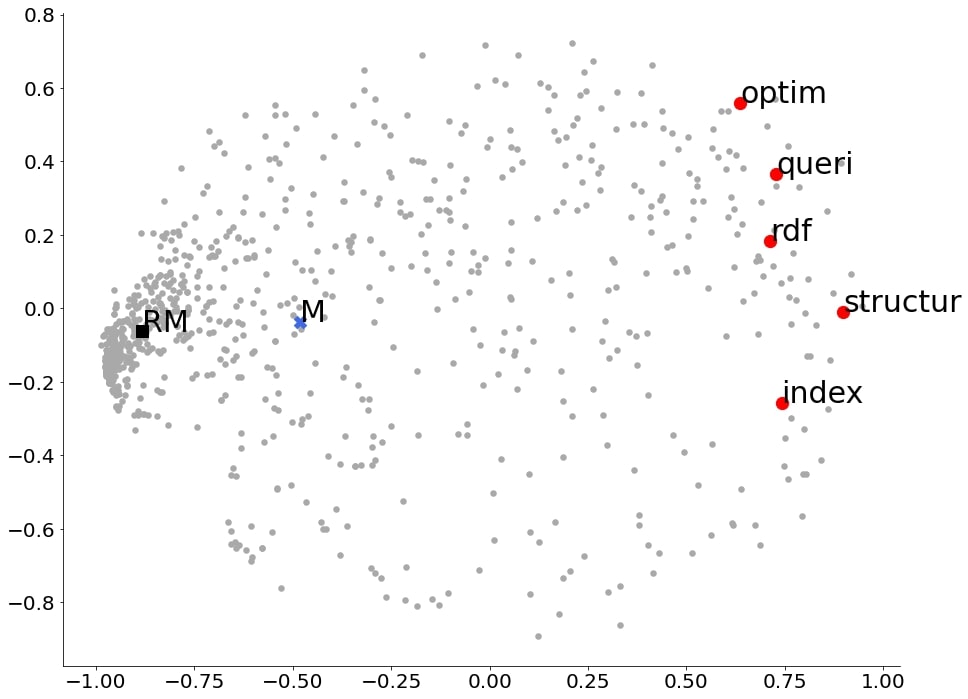}
    \caption{}
  \end{subfigure}
  
  \caption{PCA 2d projection of the normalized (a) 50d local GloVe and (b) term-term local embeddings for the article.}
  \label{fig:NUS_109_normalized_local_embeddings}
\end{figure}

Figures \ref{fig:NUS_109_unnormalized_local_embeddings} and \ref{fig:NUS_109_normalized_local_embeddings} offer insights regarding the effectiveness of the Euclidean and Mahalanobis distances compared to the cosine distance. Both figures present the PCA 2d projection on the first two principal components of the (a) 50d local GloVe and (b) term-term local embeddings for an article from the NUS collection with title ``Index structures and algorithms for querying distributed RDF repositories'' and keywords \{\textit{rdf}, \textit{structure}, \textit{optimization}, \textit{querying}, \textit{index}\}. The red circle points are the article's keywords, whereas the rest of the words are the grey ones. The original sample's center of the distribution is the blue ``x'' annotated with M (means), whereas the corresponding robust estimation is the black square annotated with RM. In particular, Fig. \ref{fig:NUS_109_normalized_local_embeddings} shows the normalized (to unit variance)\footnote{The local vectors are normalized on the $m=10$ principal components that PCA generates.} form of the vectors that the cosine distance considers, as the vectors' magnitude is out of the scope of this metric. Such normalization makes difficult the detection of keywords, as it brings closer vectors that have a similar orientation and perhaps a different magnitude (see Fig. \ref{fig:NUS_109_unnormalized_local_embeddings}), i.e., loss of valuable statistical information. Similar plots, supportive of our interpretation, are also obtained from other documents (see Figs \ref{fig:NUS_125_glove_appendix}, \ref{fig:NUS_142_glove_appendix}, \ref{fig:NUS_125_tt_appendix} and \ref{fig:NUS_142_tt_appendix} in the Appendices). 

The experimental results of Table \ref{tbl:results_trainset_custom_dist} confirm the equivalence between the two local vector representations. The local GloVe vectors is an alternative view of the t-t vectors, suitable to capture in a few vectors' dimensions the words' context, useful especially in tasks where we are interested in global vectors by training the GloVe model in a large corpus.  Hence, we suggest the usage of the t-t vector representation instead of the local GloVe one, because of its high performance in both datasets and the simplicity concerning the construction of the vectors (the GloVe model requires both term-term co-occurrences calculations and multiple runs to achieve convergence).

\subsection{Comparison with Other Approaches}
\label{subsec:comparison}

Three baselines and four state-of-the-art unsupervised AKE methods participate in this empirical study. The Tf-Idf and First-N-Words (FNW) are the main baselines of the task that utilise term frequencies in document/corpus level and the important heuristic of words' position, respectively. In particular, FNW scores the candidate unigrams according to their first occurrence in the scientific article, i.e., we consider as more significant terms those that appear earlier in the publication. At the preprocessing stage of the above baselines, we remove English stopwords, symbols, punctuation, numerals, pronouns, spaces and adverbs. We keep only unigram alphanumerics with length greater than 1. Moreover, as a third baseline, we use our approach that considers only the Euclidean distance between the distribution's center and each word vector (LV$_b$) to score the words. 

Moreover, we have experimented with the PageRank (PR) and Betweenness (BT) centrality measures that are state-of-the-art for the AKE. We also compare our method with two additional state-of-the-art methods borrowed from the keyphrase extraction task, SingleRank (SR) and PositionRank (PosR), adapted to return unigrams instead of phrases. SR, PosR and BT use only nouns and adjectives for the graph-of-words construction. However, PR builds the graph-of-words following the pre-processing steps described in Section \ref{subsub:preprocessing}. All the competitive methods run on undirected graphs with word-word co-occurrences as weights on the edges.

\begin{table}[H]
\centering
\small
\caption{F$_1$-scores of various keyword extraction methods on the test sets of 3 different datasets at the top 5, 10, 15 returned keywords.}
\begin{tabular}{|c|c|c|c|c|c|c|c|c|c|}
\hline
\multirow{2}{*}{} & \multicolumn{3}{c|}{NUS}                         & \multicolumn{3}{c|}{Semeval}                     & \multicolumn{3}{c|}{Krapivin}                         \\ \cline{2-10} 
               F$_1$   & @5             & @10            & @15            & @5             & @10            & @15            & @5             & @10            & @15            \\ \hline
Tf-Idf             & 0.273          & 0.274          & 0.255          & 0.241          & 0.268          & 0.250          & 0.196          & 0.240          & 0.244          \\ %\hline
FNW               & 0.356          & \underline{0.393}          & 0.328          & 0.312          & \textit{0.325}          & 0.285          & \underline{0.341}          & \underline{0.367}          & 0.337          \\ %\hline

SR                & \textit{0.375}          & 0.384          & \textit{0.353}          & 0.319          & 0.320          & \textit{0.295}          & 0.312          & \textit{0.353}          & \underline{0.349}          \\ %\hline
BT                & 0.361          & 0.373          & 0.346          & 0.317          & 0.309          & 0.289          & \textit{0.314}          & 0.344          & 0.335          \\ %\hline
% CLS               & 0.389          & 0.392          & 0.367          & 0.317          & 0.321          & 0.286          & 0.327          & 0.366          & 0.358          \\ %\hline
PR                & 0.373          & \textit{0.388}          & \underline{0.354}          & \textit{0.326}          & 0.318          & 0.290          & 0.306          & 0.351          & \textit{0.346}          \\ %\hline
PosR              & \underline{0.395}          & 0.387          & 0.339          & \underline{0.343}          & \underline{0.343}          & \underline{0.299}          & 0.297          & 0.335          & 0.320          \\ 
LV$_b$            & 0.354          & 0.361          & 0.315          & 0.297          & 0.299          & 0.280          & 0.277          & 0.325          & 0.323          \\ %\hline
LV               & \textbf{0.431} & \textbf{0.438} & \textbf{0.385} & \textbf{0.387} & \textbf{0.387} & \textbf{0.344} & \textbf{0.385} & \textbf{0.426} & \textbf{0.400} \\ 
\hline
\end{tabular}

\label{tbl:comparison_experiments}
\end{table}

Table \ref{tbl:comparison_experiments} shows that our approach that uses local vectors (LV) outperforms all the other baselines and state-of-the-art unsupervised methods across all datasets. PosR and FNW achieve in many cases the second-best (underlined scores) or the third-best performance (italicized scores), whereas SR and PR follow with quite high F$_1$-scores. Tf-Idf performs worse compared to the other approaches as it uses only term frequences without considering context information, while FNW seems to be a quite strong baseline of the task. The decent and comparable F$_1$-scores of LV$_b$  to the rest techniques' scores confirm the dynamics of simple local word representations to play a significant role in the AKE task. 

All the state-of-the-art methods described above as well as the LV, LV$_b$  focus on the detection of a set of words whose occurrences in a scientific publication highlight their role as keywords. The combination of term-term co-occurrence local vectors with general positional information, i.e., the id of the first sentence that a term appears (without the strict position of the first occurrence of the word), leads to the effective solution of LV. The value of the positional information is also evident from PosR and FNW that incorporate such knowledge and follow LV in the final ranking. Section \ref{subsec:qualitative_results} provides interesting insights regarding the influence of the two components, i.e., Euclidean distance and the positional information, on the final scoring scheme through qualitative results along with informative tables and figures.

\subsection{Qualitative Results}
\label{subsec:qualitative_results}

In this section, we show the positive/negative impact of the positional information on our approach. We present the extracted keywords along with their scores using the LV and LV$_b$ approaches. We also provide the information regarding each word's position via the FNW baseline. We select two scientific publications from the Semeval dataset for this study. The first article (from now on referred to as Article I) entitled as ``\textit{Live data center migration across WANs: a robust cooperative context aware approach}'' has the following set of authors' keywords \{\textit{data}, \textit{center}, \textit{migration}, \textit{virtual}, \textit{server}, \textit{storage}\}. For the second one (i.e., Article II) with the title ``\textit{Service interface: a new abstraction for implementing and composing protocols}'', authors assign as keywords the terms \{\textit{protocol}, \textit{frameworks}, \textit{modularity}, \textit{dynamic}, \textit{replacement}\}.

\begin{figure}[H]
\centering
    \centering\includegraphics[width=0.75\linewidth]{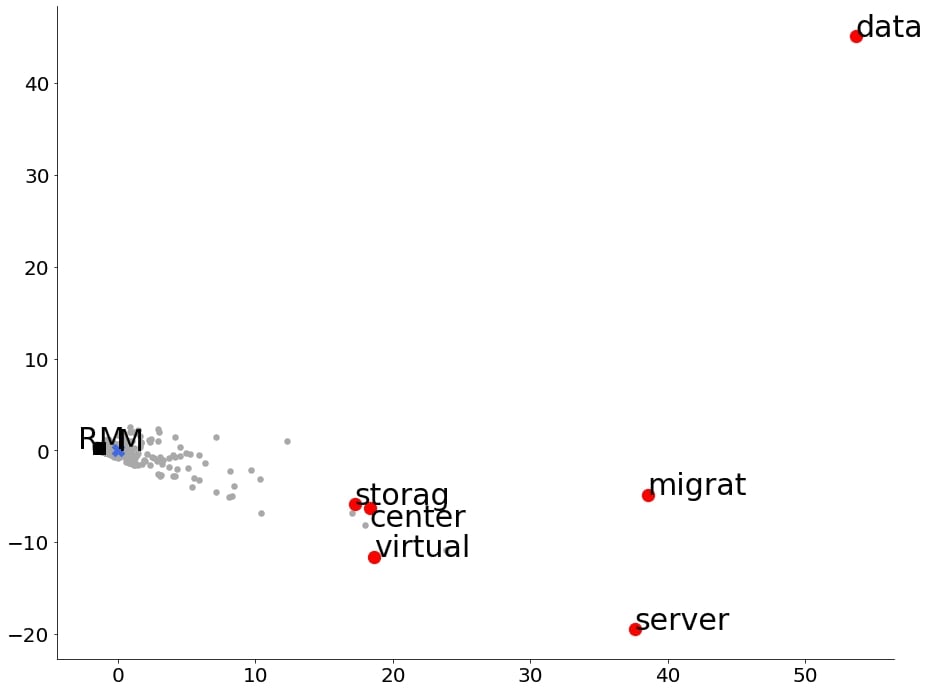}
  \caption{PCA 2d projection of the term-term local embeddings for the Article I.}
  \label{fig:qualitative_position_bad_role}
\end{figure}

Figure \ref{fig:qualitative_position_bad_role} shows the PCA 2d projection of the term-term local embeddings for the Article I. Once again, the red circle points are the authors' keywords, whereas the rest of the words are the grey ones. The original sample's center of the distribution is the blue ``x'' annotated with M, whereas the corresponding robust estimation is the black square annotated with RM. We see that this publication follows the basic assumption of the LV and LV$_b$ approaches, i.e., keywords tend to be distant from the main distribution's center. Table \ref{tbl:qualitative_neg_case} presents the returned keywords, in their stemmed form (following the evaluation approach), ranked by decreasing importance scores. The golden keywords in the output of each method are emphasised as bold. Lines also organize the results into three ordered subsets of 5 words each, to facilitate their qualitative evaluation process, i.e., easy inspection at the top 5, 10 and 15 of the ranking, respectively. The results confirm that the pure version of our approach LV$_b$ achieves to find 6/6 of the authors' keywords at the top 15 returned results. However, LV, negatively affected by the positional information, ranks lower the keywords \{\textit{server}, \textit{virtual}\}, and, finally, misses the keyword \textit{storage} (word in red bold typographical emphasis) that does not appear at the very beginning of the document.

\begin{table}[H]
\centering
\small
\caption{The top 15 returned (stemmed) keywords by LV, LV$_b$ and FWN methods, ranked by decreasing importance scores for the Article I.}
\begin{tabular}{|c|c|c|c|c|c|}
\hline
\multicolumn{2}{|c|}{\textbf{LV}} & \multicolumn{2}{c|}{\textbf{LV$_b$}}                        & \multicolumn{2}{c|}{\textbf{FWN}} \\ \hline
\textbf{data}        & 70.702     & \textbf{data}                                   & 70.702 & live                   & 0        \\ %\hline
\textbf{migrat}      & 48.888     & \textbf{migrat}                                 & 48.888 & \textbf{data}          & 1        \\ %\hline
live                 & 32.142     & \textbf{server}                                 & 48.658 & \textbf{center}        & 2        \\ %\hline
\textbf{center}      & 24.933     & live                                            & 32.142 & \textbf{migrat}        & 3        \\ %\hline
across               & 18.139     & \textbf{virtual}                                & 31.772 & wan                    & 5        \\ \hline
wan                  & 14.759     & network                                         & 30.258 & robust                 & 8        \\ %\hline
servic               & 10.329     & replic                                          & 28.099 & cooper                 & 9        \\ %\hline
\textbf{server}      & 9.732      & {\color[HTML]{FE0000} \textit{\textbf{storag}}} & 26.953 & context                & 10      \\ %\hline
approach             & 6.507      & \textbf{center}                                 & 24.933 & awar                   & 11       \\ %\hline
\textbf{virtual}     & 6.354      & applic                                          & 21.627 & approach               & 12       \\ \hline
context              & 6.202      & servic                                          & 20.657 & abstract               & 13       \\ %\hline
cooper               & 6.199      & across                                          & 18.139 & signific               & 15       \\ %\hline
avail                & 5.877      & use                                             & 16.450 & concern                & 16       \\ %\hline
replic               & 5.620      & system                                          & 15.840 & internet               & 18       \\ %\hline
outag                & 5.494      & remot                                           & 15.753 & base                   & 20       \\ \hline
\end{tabular}

\label{tbl:qualitative_neg_case}
\end{table}

\begin{figure}[H]
\centering
  
    \centering\includegraphics[width=0.75\linewidth]{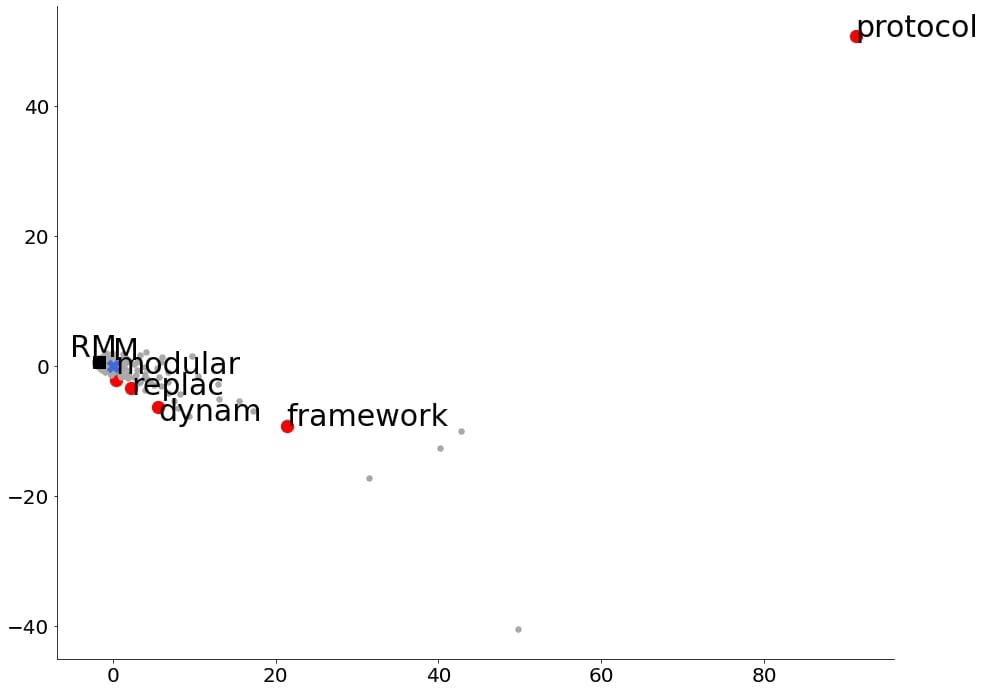}
 
  \caption{PCA 2d projection of the term-term local embeddings for the Article II.}
  \label{fig:qualitative_position_good_role}
\end{figure}

On the contrary, Fig. \ref{fig:qualitative_position_good_role} shows that the keywords of Article II are not all quite distant from the main distribution's center. Again, Table \ref{tbl:qualitative_pos_case} presents the returned keywords ranked by decreasing importance. In this case, LV$_b$ performs poorly, failing to detect the 50\% (2/4) of the authors' keywords at the top 15 returned results. However, LV, positively affected by the positional information, although it wrongly ranks lower the keywords \{\textit{protocol}, \textit{framework}\}, finally, manages to detect the keyword \textit{dynamic} (word in green bold typographical emphasis) that appears earlier in the document compared to the words (stemmed here) \textit{request}, \textit{stack}, \textit{bound}, etc.

\begin{table}[H]
\centering
\small
\caption{The top 15 returned (stemmed) keywords by LV, LV$_b$ and FWN methods, ranked by decreasing importance scores for the Article II.}
\begin{tabular}{|c|c|c|c|c|c|}
\hline
\multicolumn{2}{|c|}{\textbf{LV}}              & \multicolumn{2}{c|}{\textbf{LV$_b$}} & \multicolumn{2}{c|}{\textbf{FNW}} \\ \hline
servic                                & 73.617 & \textbf{protocol}     & 104.887   & servic                  & 0       \\ %\hline
\textbf{protocol}                     & 52.444 & servic                & 73.617    & interfac                & 1       \\ %\hline
modul                                 & 24.997 & base                  & 57.618    & abstract                & 5       \\ %\hline
interfac                              & 12.685  & event                 & 53.070    & implement               & 7       \\ %\hline
implement                             & 10.614 & modul                 & 48.862    & compos                  & 9       \\ \hline
event                                 & 9.772  & \textbf{framework}    & 30.390    & \textbf{protocol}       & 10      \\ %\hline
base                                  & 9.603  & implement             & 25.370    & paper                   & 15      \\ %\hline
\textbf{framework}                    & 7.597  & interfac              & 24.997    & compar                  & 17      \\ %\hline
compos                                & 3.898  & request               & 24.821    & approach                & 19      \\ %\hline
approach                              & 2.996  & execut                & 22.533    & design                  & 22      \\ \hline
network                               & 2.713  & bound                 & 20.544    & \textbf{framework}      & 25      \\ %\hline
abstract                              & 2.559  & issu                  & 19.908    & tool                    & 27      \\ %\hline
interact                              & 2.332  & stack                 & 19.366    & \textbf{modular}                 & 30      \\ %\hline
{\color[HTML]{009901} \textit{\textbf{dynam}}} & 2.191  & handl                 & 18.090    & network                 & 31      \\ %\hline
compar                                & 1.488  & notif                 & 17.331    & common                  & 36      \\ \hline
\end{tabular}

\label{tbl:qualitative_pos_case}
\end{table}

\section{Conclusions and Future Work}
\label{sec:conclusions}

This work presents a new unsupervised keyword extraction approach, whose main innovation is that we consider the distribution's center closer to the non-keywords, since the main bulk of words are neutral or slightly relevant to the documents' topics, in contrast to existing methods. Our empirical study offers evidence that local word vector representations, such as GloVe or equivalent/simpler local vectors based on the words' co-occurrences, can lead to better keyword extraction results compared to popular state-of-the-art unsupervised approaches. Unlike the main direction in text mining that uses the cosine similarity/distance, we recommend distance metrics that incorporate the vectors' magnitude (Euclidean distance) or covariance estimations (Mahalanobis distance) to capture the similar behaviour of the non-keywords' vectors over the keywords' ones.

We hope this work paves the way to the researchers for further investigation on local representations. In particular, we envisage implications of our work towards (a) improved keyword extraction methods based on local word embeddings, (b) applying local word embeddings to other information processing tasks, and (c) developing novel methods for learning local word embeddings that will incorporate positional information or even semantics. Soon, we intend to build on top of this work, a solution that estimates the effectiveness of robust estimations regarding the distribution's center, covariance matrix, etc. based on the properties of the document's local vector representations.

\bibliography{Mendeley_alma}

\begin{thebibliography}{}

\bibitem[Barrat et~al., 2004]{barrat2004architecture}
{\bf Barrat, A.}, {\bf Barthelemy, M.}, {\bf Pastor-Satorras, R.}, \textbf{and}
  {\bf Vespignani, A.} 2004.
\newblock The architecture of complex weighted networks.
\newblock {\em Proceedings of the National Academy of Sciences},
  101(11):3747--3752.

\bibitem[Batagelj and Zaversnik, 2002]{DBLP:journals/corr/cs-DS-0202039}
{\bf Batagelj, V.} \textbf{and} {\bf Zaversnik, M.} 2002.
\newblock Generalized cores.
\newblock {\em CoRR}, cs.DS/0202039.

\bibitem[Batagelj and Zaversnik, 2003]{DBLP:journals/corr/cs-DS-0310049}
{\bf Batagelj, V.} \textbf{and} {\bf Zaversnik, M.} 2003.
\newblock An {O}(m) algorithm for cores decomposition of networks.
\newblock {\em CoRR}, cs.DS/0310049.

\bibitem[Bennani{-}Smires et~al., 2018]{DBLP:conf/conll/Bennani-SmiresM18}
{\bf Bennani{-}Smires, K.}, {\bf Musat, C.}, {\bf Hossmann, A.}, {\bf
  Baeriswyl, M.}, \textbf{and} {\bf Jaggi, M.} 2018.
\newblock Simple unsupervised keyphrase extraction using sentence embeddings.
\newblock In {\em Proceedings of the 22nd Conference on Computational Natural
  Language Learning, CoNLL 2018}, pp. 221--229, Brussels, Belgium. Association
  for Computational Linguistics.

\bibitem[Biswas et~al., 2018]{DBLP:journals/eswa/BiswasBS18}
{\bf Biswas, S.~K.}, {\bf Bordoloi, M.}, \textbf{and} {\bf Shreya, J.} 2018.
\newblock A graph based keyword extraction model using collective node weight.
\newblock {\em Expert Systems with Applications}, 97:51--59.

\bibitem[Bojanowski et~al., 2017]{DBLP:journals/tacl/BojanowskiGJM17}
{\bf Bojanowski, P.}, {\bf Grave, E.}, {\bf Joulin, A.}, \textbf{and} {\bf
  Mikolov, T.} 2017.
\newblock Enriching word vectors with subword information.
\newblock {\em Transactions of the Association for Computational Linguistics},
  5:135--146.

\bibitem[Brin and Page, 1998]{grin+page1998}
{\bf Brin, S.} \textbf{and} {\bf Page, L.} 1998.
\newblock The anatomy of a large-scale hypertextual web search engine.
\newblock {\em Computer Networks}, 30(1-7):107--117.

\bibitem[Campos et~al., 2020]{DBLP:journals/isci/CamposMPJNJ20}
{\bf Campos, R.}, {\bf Mangaravite, V.}, {\bf Pasquali, A.}, {\bf Jorge, A.},
  {\bf Nunes, C.}, \textbf{and} {\bf Jatowt, A.} 2020.
\newblock Yake! keyword extraction from single documents using multiple local
  features.
\newblock {\em Information Sciences Journal}, 509:257--289.

\bibitem[Campos et~al., 2018]{DBLP:conf/ecir/0001MPJNJ18}
{\bf Campos, R.}, {\bf Mangaravite, V.}, {\bf Pasquali, A.}, {\bf Jorge,
  A.~M.}, {\bf Nunes, C.}, \textbf{and} {\bf Jatowt, A.} 2018.
\newblock A text feature based automatic keyword extraction method for single
  documents.
\newblock In {\em Advances in Information Retrieval - Proceedings of the 40th
  European Conference on {IR} Research, {ECIR} 2018}, volume 10772 of {\em
  Lecture Notes in Computer Science}, pp. 684--691, Grenoble, France. Springer.

\bibitem[Cohen, 2008]{cohen2008trusses}
{\bf Cohen, J.} 2008.
\newblock Trusses: Cohesive subgraphs for social network analysis.
\newblock In {\em Technical Report}, volume~16, pp. 3--29. National Security
  Agency.

\bibitem[Collobert et~al., 2011]{DBLP:journals/jmlr/CollobertWBKKK11}
{\bf Collobert, R.}, {\bf Weston, J.}, {\bf Bottou, L.}, {\bf Karlen, M.}, {\bf
  Kavukcuoglu, K.}, \textbf{and} {\bf Kuksa, P.~P.} 2011.
\newblock Natural language processing (almost) from scratch.
\newblock {\em Journal of Machine Learning Research}, 12:2493--2537.

\bibitem[Florescu and Caragea, 2017]{DBLP:conf/acl/FlorescuC17}
{\bf Florescu, C.} \textbf{and} {\bf Caragea, C.} 2017.
\newblock Position{R}ank: An unsupervised approach to keyphrase extraction from
  scholarly documents.
\newblock In {\em Proceedings of the 55th Annual Meeting of the Association for
  Computational Linguistics, {ACL} 2017}, pp. 1105--1115, Vancouver, Canada.

\bibitem[Gutwin et~al., 1999]{gutwin1999improving}
{\bf Gutwin, C.}, {\bf Paynter, G.~W.}, {\bf Witten, I.~H.}, {\bf
  Nevill{-}Manning, C.~G.}, \textbf{and} {\bf Frank, E.} 1999.
\newblock Improving browsing in digital libraries with keyphrase indexes.
\newblock {\em Decision Support Systems}, 27(1-2):81--104.

\bibitem[Hasan and Ng, 2014]{hasan+ng2014}
{\bf Hasan, K.~S.} \textbf{and} {\bf Ng, V.} 2014.
\newblock Automatic keyphrase extraction: {A} survey of the state of the art.
\newblock In {\em Proceedings of the 52nd Annual Meeting of the Association for
  Computational Linguistics, {ACL} 2014, (Volume 1: Long Papers)}, pp.
  1262--1273, Baltimore, MD, USA.

\bibitem[Hotelling, 1933]{hotelling1933analysis}
{\bf Hotelling, H.} 1933.
\newblock Analysis of a complex of statistical variables into principal
  components.
\newblock {\em Journal of educational psychology}, 24(6):417--441.

\bibitem[Hulth and Megyesi, 2006]{DBLP:conf/acl/HulthM06}
{\bf Hulth, A.} \textbf{and} {\bf Megyesi, B.} 2006.
\newblock A study on automatically extracted keywords in text categorization.
\newblock In {\em Proceedings of the 21st International Conference on
  Computational Linguistics and the 44th Annual Meeting of the Association for
  Computational Linguistics, {ACL} 2006}, pp. 537--544, Sydney, Australia. The
  Association for Computational Linguistics.

\bibitem[Joulin et~al., 2017]{DBLP:conf/eacl/GraveMJB17}
{\bf Joulin, A.}, {\bf Grave, E.}, {\bf Bojanowski, P.}, \textbf{and} {\bf
  Mikolov, T.} 2017.
\newblock Bag of tricks for efficient text classification.
\newblock In {\em Proceedings of the 15th Conference of the European Chapter of
  the Association for Computational Linguistics, {EACL} 2017, Volume 2: Short
  Papers}, pp. 427--431, Valencia, Spain. The Association for Computational
  Linguistics.

\bibitem[Kim et~al., 2010]{Kim:semeval2010}
{\bf Kim, S.~N.}, {\bf Medelyan, O.}, {\bf Kan, M.}, \textbf{and} {\bf Baldwin,
  T.} 2010.
\newblock Semeval-2010 task 5 : Automatic keyphrase extraction from scientific
  articles.
\newblock In {\em Proceedings of the 5th International Workshop on Semantic
  Evaluation, SemEval@ACL 2010}, pp. 21--26, Uppsala, Sweden.

\bibitem[Krapivin et~al., 2008]{krapivin2009}
{\bf Krapivin, M.}, {\bf Autayeu, A.}, \textbf{and} {\bf Marchese, M.} 2008.
\newblock Large dataset for keyphrases extraction.
\newblock In {\em Technical Report DISI-09-055}. Trento, Italy.

\bibitem[Lahiri et~al., 2014]{DBLP:journals/corr/LahiriCC14}
{\bf Lahiri, S.}, {\bf Choudhury, S.~R.}, \textbf{and} {\bf Caragea, C.} 2014.
\newblock Keyword and keyphrase extraction using centrality measures on
  collocation networks.
\newblock {\em CoRR}, abs/1401.6571.

\bibitem[Lau and Baldwin, 2016]{DBLP:conf/rep4nlp/LauB16}
{\bf Lau, J.~H.} \textbf{and} {\bf Baldwin, T.} 2016.
\newblock An empirical evaluation of doc2vec with practical insights into
  document embedding generation.
\newblock In {\em Proceedings of the 1st Workshop on Representation Learning
  for NLP, Rep4NLP@ACL 2016}, pp. 78--86, Berlin, Germany. The Association for
  Computational Linguistics.

\bibitem[Mahata et~al., 2018]{DBLP:conf/naacl/MahataKSZ18}
{\bf Mahata, D.}, {\bf Kuriakose, J.}, {\bf Shah, R.~R.}, \textbf{and} {\bf
  Zimmermann, R.} 2018.
\newblock Key2vec: Automatic ranked keyphrase extraction from scientific
  articles using phrase embeddings.
\newblock In {\em Proceedings of the 2018 Conference of the North American
  Chapter of the Association for Computational Linguistics: Human Language
  Technologies, NAACL-HLT 2018, Volume 2 (Short Papers)}, pp. 634--639, New
  Orleans, Louisiana, USA. The Association for Computational Linguistics.

\bibitem[Medelyan et~al., 2009]{medelyan2009human}
{\bf Medelyan, O.}, {\bf Frank, E.}, \textbf{and} {\bf Witten, I.~H.} 2009.
\newblock Human-competitive tagging using automatic keyphrase extraction.
\newblock In {\em Proceedings of the 2009 Conference on Empirical Methods in
  Natural Language Processing, {EMNLP} 2009}, pp. 1318--1327, Singapore.

\bibitem[Meng et~al., 2017]{DBLP:conf/acl/MengZHHBC17}
{\bf Meng, R.}, {\bf Zhao, S.}, {\bf Han, S.}, {\bf He, D.}, {\bf Brusilovsky,
  P.}, \textbf{and} {\bf Chi, Y.} 2017.
\newblock Deep keyphrase generation.
\newblock In {\em Proceedings of the 55th Annual Meeting of the Association for
  Computational Linguistics, {ACL} 2017, Volume 1: Long Papers}, pp. 582--592,
  Vancouver, Canada. The Association for Computational Linguistics.

\bibitem[Merrouni et~al., 2019]{DBLP:journals/jiis/MerrouniFO20}
{\bf Merrouni, Z.~A.}, {\bf Frikh, B.}, \textbf{and} {\bf Ouhbi, B.} 2019.
\newblock Automatic keyphrase extraction: A survey and trends.
\newblock {\em Journal of Intelligent Information Systems}, 54(2):391--424.

\bibitem[Mihalcea and Tarau, 2004]{mihalcea+tatau2004}
{\bf Mihalcea, R.} \textbf{and} {\bf Tarau, P.} 2004.
\newblock Text{R}ank: Bringing order into text.
\newblock In {\em Proceedings of the 2004 Conference on Empirical Methods in
  Natural Language Processing, {EMNLP} 2004}, pp. 404--411, Barcelona, Spain.

\bibitem[Mikolov et~al., 2013]{DBLP:journals/corr/abs-1301-3781}
{\bf Mikolov, T.}, {\bf Chen, K.}, {\bf Corrado, G.}, \textbf{and} {\bf Dean,
  J.} 2013.
\newblock Efficient estimation of word representations in vector space.
\newblock In {\em Proceedings of the 2013 International Conference on Learning
  Representations, ICLR 2013, Workshop Track}, Scottsdale, Arizona, USA.

\bibitem[Mnih and Hinton, 2007]{DBLP:conf/icml/MnihH07}
{\bf Mnih, A.} \textbf{and} {\bf Hinton, G.~E.} 2007.
\newblock Three new graphical models for statistical language modelling.
\newblock In {\em Proceedings of the Twenty-Fourth International Conference on
  Machine Learning, {(ICML} 2007)}, volume 227 of {\em {ACM} International
  Conference Proceeding Series}, pp. 641--648, Corvallis, Oregon, USA. {ACM}.

\bibitem[Nguyen and Kan, 2007]{DBLP:conf/icadl/NguyenK07}
{\bf Nguyen, T.~D.} \textbf{and} {\bf Kan, M.} 2007.
\newblock Keyphrase extraction in scientific publications.
\newblock In {\em Proceedings of the Asian Digital Libraries. Looking Back 10
  Years and Forging New Frontiers, 10th International Conference on Asian
  Digital Libraries, {ICADL} 2007}, volume 4822 of {\em Lecture Notes in
  Computer Science}, pp. 317--326, Hanoi, Vietnam. Springer.

\bibitem[Pagliardini et~al., 2018]{DBLP:conf/naacl/PagliardiniGJ18}
{\bf Pagliardini, M.}, {\bf Gupta, P.}, \textbf{and} {\bf Jaggi, M.} 2018.
\newblock Unsupervised learning of sentence embeddings using compositional
  n-gram features.
\newblock In {\em Proceedings of the 2018 Conference of the North American
  Chapter of the Association for Computational Linguistics: Human Language
  Technologies, {NAACL-HLT} 2018, Volume 1 (Long Papers)}, pp. 528--540, New
  Orleans, Louisiana, USA. The Association for Computational Linguistics.

\bibitem[Papagiannopoulou and Tsoumakas, 2018]{papagiannopoulou2018local}
{\bf Papagiannopoulou, E.} \textbf{and} {\bf Tsoumakas, G.} 2018.
\newblock Local word vectors guiding keyphrase extraction.
\newblock {\em Information Processing \& Management}, 54(6):888--902.

\bibitem[Papagiannopoulou and Tsoumakas, 2019]{Papagiannopoulou2019outliers}
{\bf Papagiannopoulou, E.} \textbf{and} {\bf Tsoumakas, G.} 2019.
\newblock Unsupervised keyphrase extraction from scientific publications.
\newblock {\em To appear in Proceedings of the 20th International Conference on
  Computational Linguistics and Intelligent Text Processing, CICLing 2019}, La
  Rochelle, France.

\bibitem[Papagiannopoulou and Tsoumakas, 2020]{papagiannopoulou2019review}
{\bf Papagiannopoulou, E.} \textbf{and} {\bf Tsoumakas, G.} 2020.
\newblock A review of keyphrase extraction.
\newblock {\em Wiley Interdisciplinary Reviews (WIREs): Data Mining and
  Knowledge Discovery}, 10(2).

\bibitem[Pearson, 1901]{pearson1901liii}
{\bf Pearson, K.} 1901.
\newblock {LIII}. {O}n lines and planes of closest fit to systems of points in
  space.
\newblock {\em The London, Edinburgh, and Dublin Philosophical Magazine and
  Journal of Science}, 2(11):559--572.

\bibitem[Pennington et~al., 2014]{Pennington14glove:global}
{\bf Pennington, J.}, {\bf Socher, R.}, \textbf{and} {\bf Manning, C.~D.} 2014.
\newblock Glove: Global vectors for word representation.
\newblock In {\em Proceedings of the 2014 Conference on Empirical Methods in
  Natural Language Processing, {EMNLP} 2014}, pp. 1532--1543, Doha, Qatar. The
  Association for Computational Linguistics.

\bibitem[Rose et~al., 2010]{rose2010automatic}
{\bf Rose, S.}, {\bf Engel, D.}, {\bf Cramer, N.}, \textbf{and} {\bf Cowley,
  W.} 2010.
\newblock Automatic keyword extraction from individual documents.
\newblock {\em Text mining: applications and theory}, 1:1--20.

\bibitem[Rossi, 2018]{rossi2018mathematical}
{\bf Rossi, R.~J.} 2018.
\newblock {\em Mathematical statistics: An introduction to likelihood based
  inference}.
\newblock John Wiley \& Sons.

\bibitem[Rousseau and Vazirgiannis, 2015]{DBLP:conf/ecir/RousseauV15}
{\bf Rousseau, F.} \textbf{and} {\bf Vazirgiannis, M.} 2015.
\newblock Main core retention on graph-of-words for single-document keyword
  extraction.
\newblock In {\em Proceedings of the Advances in Information Retrieval - 37th
  European Conference on {IR} Research, {ECIR} 2015}, pp. 382--393, Vienna,
  Austria.

\bibitem[Rousseeuw, 1984]{Rousseeuw1984LeastRegression}
{\bf Rousseeuw, P.~J.} 1984.
\newblock {Least median of squares regression}.
\newblock {\em Journal of the American Statistical Association},
  79(388):871--880.

\bibitem[Rousseeuw and van Driessen,
  1999]{DBLP:journals/technometrics/RousseeuwD99}
{\bf Rousseeuw, P.~J.} \textbf{and} {\bf van Driessen, K.} 1999.
\newblock A fast algorithm for the minimum covariance determinant estimator.
\newblock {\em Technometrics}, 41(3):212--223.

\bibitem[Seidman, 1983]{seidman1983network}
{\bf Seidman, S.~B.} 1983.
\newblock Network structure and minimum degree.
\newblock {\em Social networks}, 5(3):269--287.

\bibitem[Song et~al., 2006]{DBLP:conf/jcdl/SongSAO06}
{\bf Song, M.}, {\bf Song, I.}, {\bf Allen, R.~B.}, \textbf{and} {\bf
  Obradovic, Z.} 2006.
\newblock Keyphrase extraction-based query expansion in digital libraries.
\newblock In {\em Proceedings of the {ACM/IEEE} Joint Conference on Digital
  Libraries, {JCDL} 2006}, pp. 202--209, Chapel Hill, NC, USA. {ACM}.

\bibitem[Tixier et~al., 2016]{DBLP:conf/emnlp/TixierMV16}
{\bf Tixier, A.~J.}, {\bf Malliaros, F.~D.}, \textbf{and} {\bf Vazirgiannis,
  M.} 2016.
\newblock A graph degeneracy-based approach to keyword extraction.
\newblock In {\em Proceedings of the 2016 Conference on Empirical Methods in
  Natural Language Processing, {EMNLP} 2016}, pp. 1860--1870, Austin, Texas,
  USA. The Association for Computational Linguistics.

\bibitem[Turian et~al., 2010]{DBLP:conf/acl/TurianRB10}
{\bf Turian, J.~P.}, {\bf Ratinov, L.}, \textbf{and} {\bf Bengio, Y.} 2010.
\newblock Word representations: {A} simple and general method for
  semi-supervised learning.
\newblock In {\em Proceedings of the 48th Annual Meeting of the Association for
  Computational Linguistics, {ACL} 2010}, pp. 384--394, Uppsala, Sweden. The
  Association for Computer Linguistics.

\bibitem[Vega{-}Oliveros et~al., 2019]{DBLP:journals/ipm/Vega-OliverosGM19}
{\bf Vega{-}Oliveros, D.~A.}, {\bf Gomes, P.~S.}, {\bf Milios, E.~E.},
  \textbf{and} {\bf Berton, L.} 2019.
\newblock A multi-centrality index for graph-based keyword extraction.
\newblock {\em Information Processing \& Management}, 56(6).

\bibitem[Wan and Xiao, 2008]{wan+xiao2008}
{\bf Wan, X.} \textbf{and} {\bf Xiao, J.} 2008.
\newblock Single document keyphrase extraction using neighborhood knowledge.
\newblock In {\em Proceedings of the 23rd {AAAI} Conference on Artificial
  Intelligence, {AAAI} 2008}, pp. 855--860, Chicago, Illinois, USA. {AAAI}
  Press.

\bibitem[Wang et~al., 2014]{Wang2014}
{\bf Wang, R.}, {\bf Liu, W.}, \textbf{and} {\bf McDonald, C.} 2014.
\newblock Corpus-independent generic keyphrase extraction using word embedding
  vectors.
\newblock In {\em Proceedings of the Software Engineering Research Conference},
  volume~39, pp. 1--8.

\bibitem[Wang et~al., 2015]{DBLP:conf/adc/WangLM15}
{\bf Wang, R.}, {\bf Liu, W.}, \textbf{and} {\bf McDonald, C.} 2015.
\newblock Using word embeddings to enhance keyword identification for
  scientific publications.
\newblock In {\em Proceedings of the Databases Theory and Applications - 26th
  Australasian Database Conference, {ADC} 2015}, pp. 257--268, Melbourne, VIC,
  Australia.

\bibitem[Zhang et~al., 2004]{DBLP:journals/wias/ZhangZM04}
{\bf Zhang, Y.}, {\bf Zincir{-}Heywood, A.~N.}, \textbf{and} {\bf Milios,
  E.~E.} 2004.
\newblock World {W}ide {W}eb site summarization.
\newblock {\em Web Intelligence and Agent Systems: An International Journal},
  2(1):39--53.

\end{thebibliography}

\label{lastpage}

\newpage

\appendix
\label{app:appendix}

\section{Local GloVe Word Vectors}
\label{app:glove}

\begin{figure}[H]
\centering

  \begin{subfigure}{0.45\linewidth}
    \centering\includegraphics[width=1.0\linewidth]{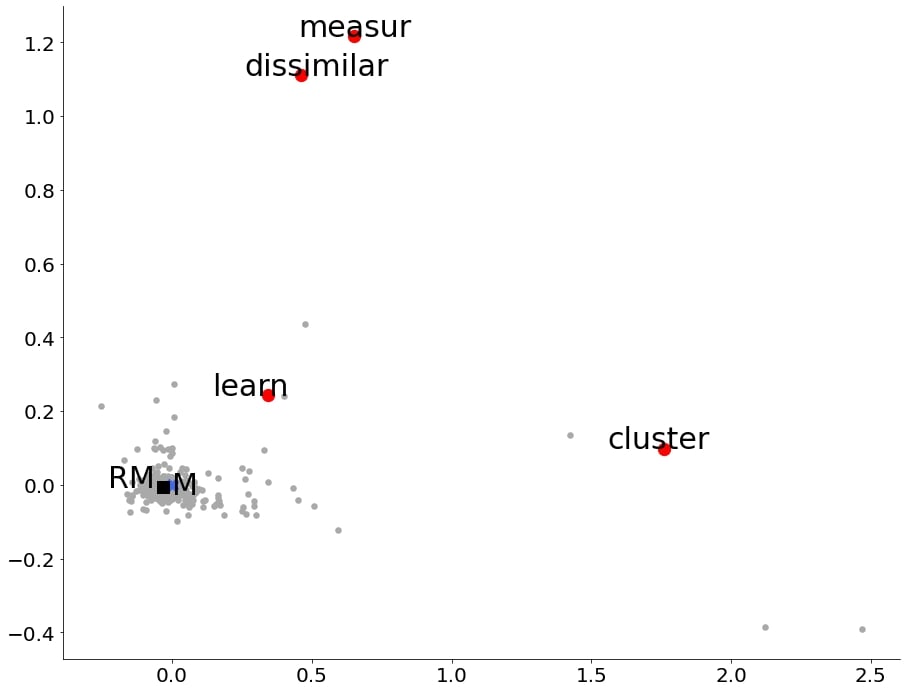}
    \caption{}
  \end{subfigure}
  \begin{subfigure}{0.45\linewidth}
    \centering\includegraphics[width=1.0\linewidth]{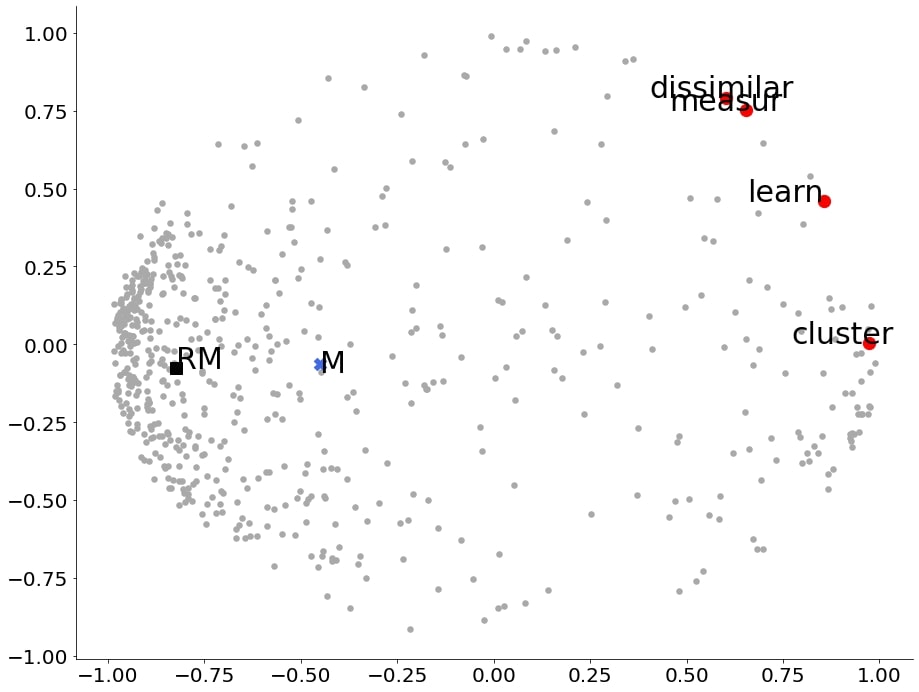}
    \caption{}
  \end{subfigure}
  \caption{(a) unnormalized and (b) normalized PCA 2d projection of the  GloVe vectors for the article with title ``Learning Spatially Variant Dissimilarity (SVaD) measures'' and keywords \{\textit{learning}, \textit{dissimilarity}, \textit{measures}, \textit{clustering}\}.}
  \label{fig:NUS_125_glove_appendix}
\end{figure}

\begin{figure}[h]
\centering
  \begin{subfigure}{0.45\linewidth}
    \centering\includegraphics[width=1.0\linewidth]{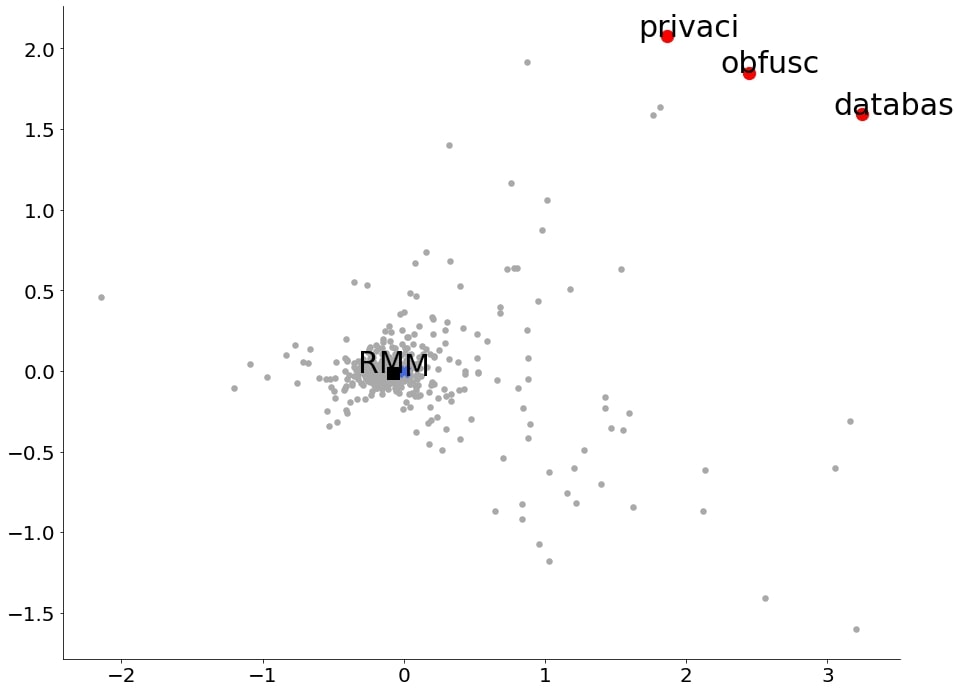}
    \caption{}
  \end{subfigure}
  \begin{subfigure}{0.45\linewidth}
    \centering\includegraphics[width=1.0\linewidth]{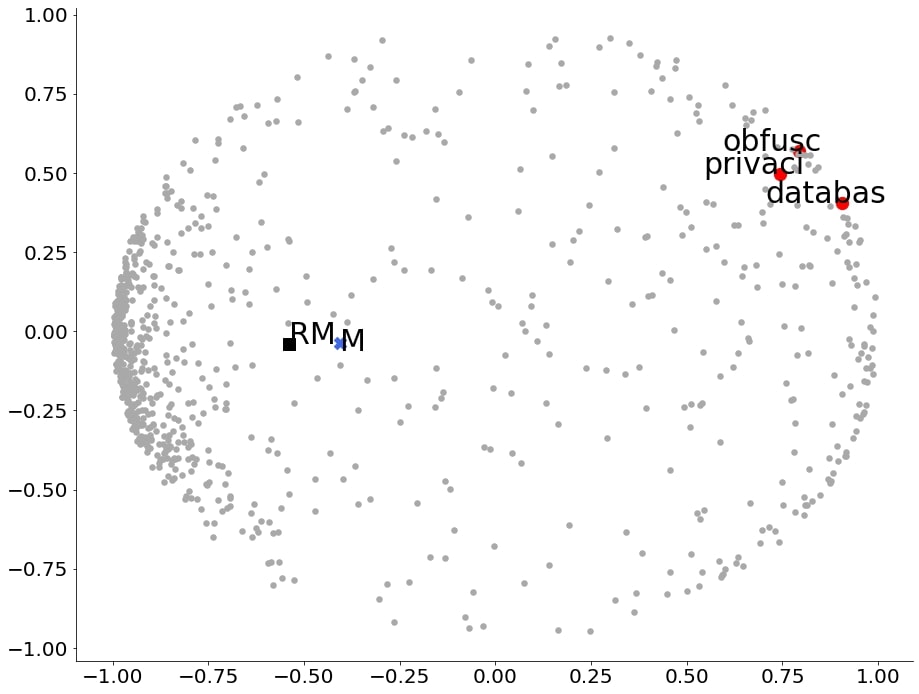}
    \caption{}
  \end{subfigure}
 
  \caption{(a) unnormalized and (b) normalized PCA 2d projection of the  GloVe vectors for the article with title ``Obfuscated databases and group privacy'' and keywords \{\textit{obfuscation}, \textit{database}, \textit{privacy}\}.}
  \label{fig:NUS_142_glove_appendix}
\end{figure}

\begin{figure}[H]
\centering
  
    \centering
    
    \includegraphics[width=0.75\linewidth]{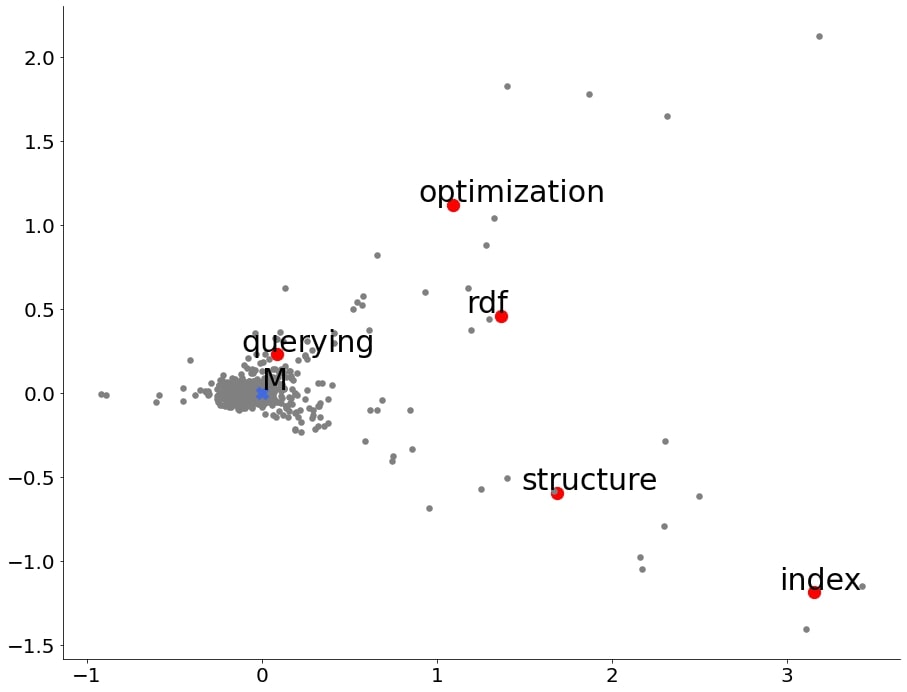}
  
  \caption{PCA 2d projection of the 50d local GloVe vectors for the article from the NUS collection with title ``Index structures and algorithms for querying distributed RDF repositories'' and keywords \{\textit{rdf}, \textit{structure}, \textit{optimization}, \textit{querying}, \textit{index}\}.}
  \label{fig:inspiration2}
\end{figure}

\section{Local Vectors based on Term-Term Co-occurrences}
\label{app:t-t}

\begin{figure}[H]
\centering

  \begin{subfigure}{0.45\linewidth}
    \centering\includegraphics[width=1.0\linewidth]{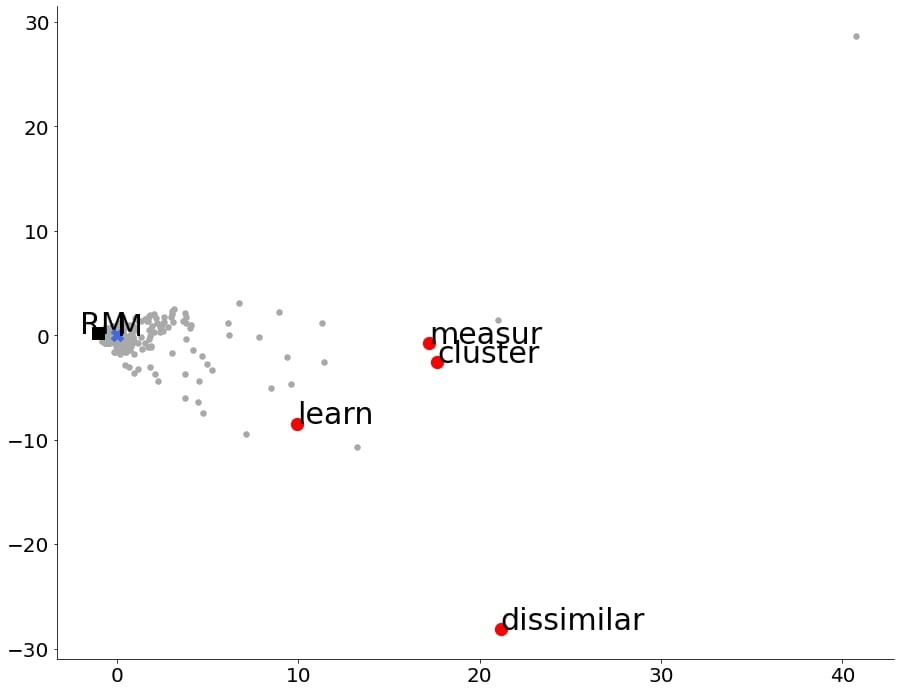}
    \caption{}
  \end{subfigure}
  \begin{subfigure}{0.45\linewidth}
    \centering\includegraphics[width=1.0\linewidth]{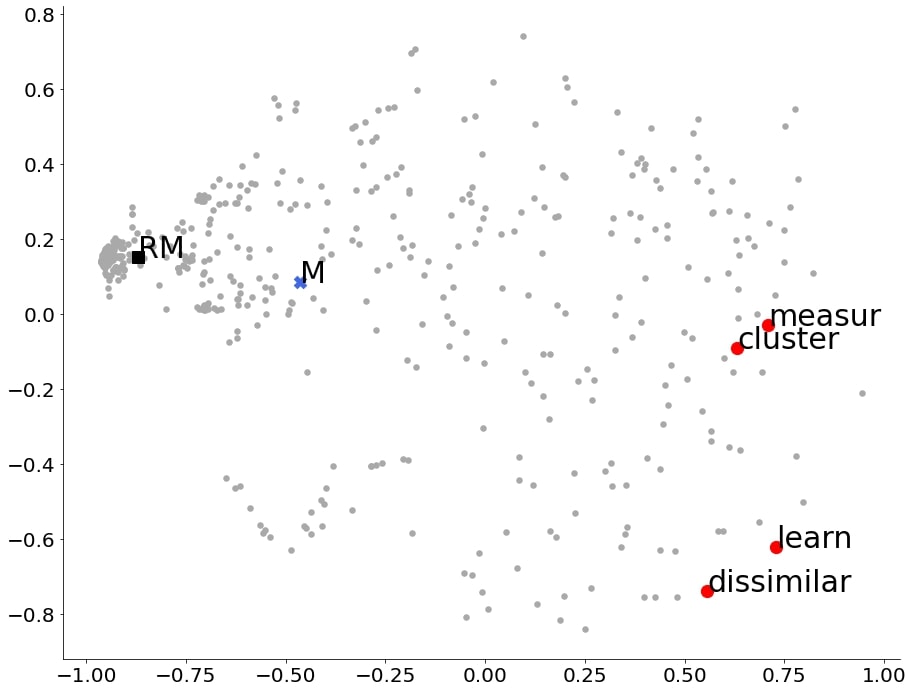}
    \caption{}
  \end{subfigure}
    \caption{(a) unnormalized and (b) normalized PCA 2d projection of the  term-term vectors for the article with title ``Learning Spatially Variant Dissimilarity (SVaD) measures'' and keywords  \{\textit{learning}, \textit{dissimilarity}, \textit{measures}, \textit{clustering}\}.}
  \label{fig:NUS_125_tt_appendix}
\end{figure}

\begin{figure}[H]
\centering
  \begin{subfigure}{0.45\linewidth}
    \centering\includegraphics[width=1.0\linewidth]{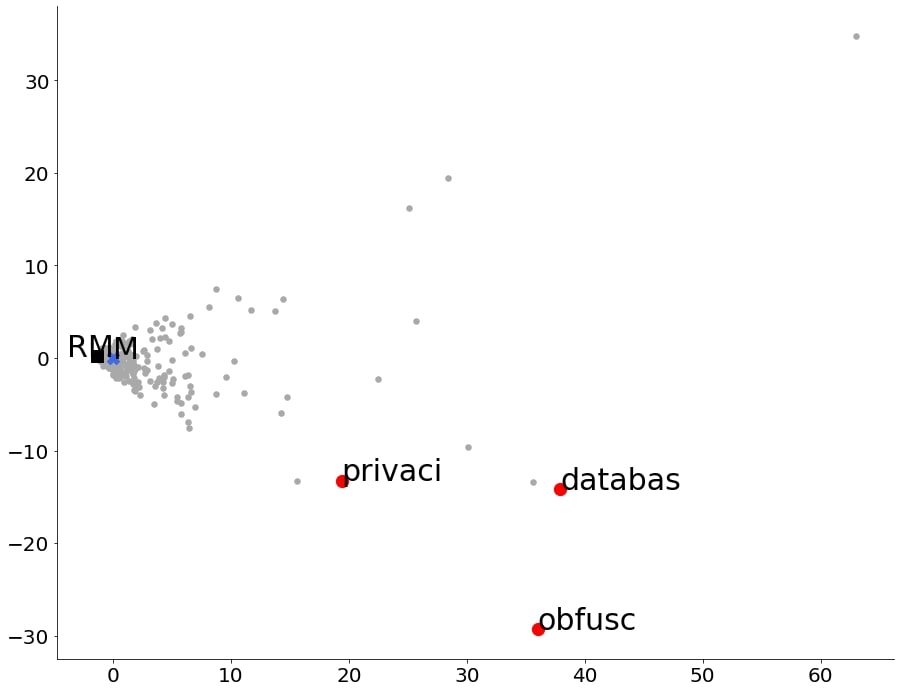}
    \caption{}
  \end{subfigure}
  \begin{subfigure}{0.45\linewidth}
    \centering\includegraphics[width=1.0\linewidth]{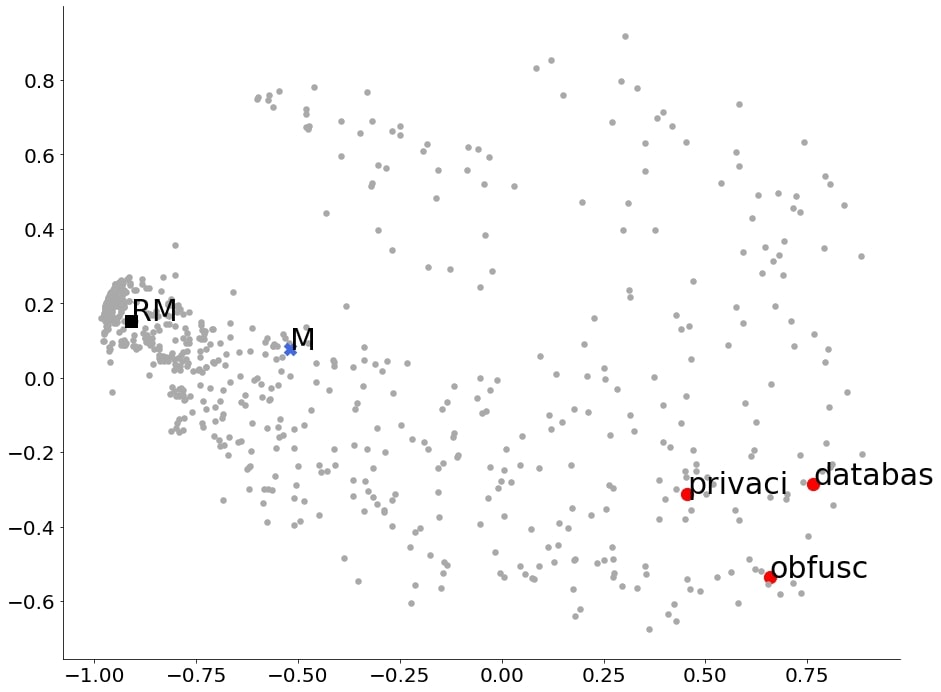}
    \caption{}
  \end{subfigure}
 
  \caption{(a) unnormalized and (b) normalized PCA 2d projection of the  term-term vectors for the article with title ``Obfuscated databases and group privacy'' and keywords \{\textit{obfuscation}, \textit{database}, \textit{privacy}\}.}
  \label{fig:NUS_142_tt_appendix}
\end{figure}

\end{document}